\documentclass[11pt,letterpaper]{article}
\usepackage{ijcnlp2017}
\usepackage{times}
\usepackage{latexsym}
\usepackage{graphicx}
\usepackage{arydshln}
\usepackage{xspace}
\usepackage{adjustbox}
\usepackage{multirow}
\usepackage{amsmath}
\usepackage{amsfonts}
\usepackage{amssymb}
\usepackage{url}
\usepackage{subcaption}
\usepackage{soul}
\usepackage{relsize}

\ijcnlpfinalcopy
\newcommand{\method}[1]{\texttt{#1}\xspace}
\newcommand{\deepgeo}{\method{deepgeo}}
\newcommand{\deepgeon}{\method{deepgeo$+$noise}}
\newcommand{\deepgeol}{\method{deepgeo$+$loss}}
\newcommand{\lsh}{\method{lsh}}
\newcommand{\fh}{\method{fasthash}}
\newcommand{\wv}{\method{word2vec}}
\newcommand{\sg}{\method{skip-gram}}
\newcommand{\fasttext}{\method{fastText}}

\newcommand{\ex}[1]{\textit{#1}}
\newcommand{\mn}[2]{\multirow{#1}{*}{\texttt{#2}}}
\newcommand{\tl}[2]{\multirow{#1}{*}{\begin{tabular}[c]{@{}c@{}}#2\end{tabular}}}

\newcommand{\secref}[2][]{Section#1~\ref{sec:#2}}

\newcommand{\tabref}[2][]{Table#1~\ref{tab:#2}}
\newcommand{\figref}[2][]{Figure#1~\ref{fig:#2}}

\newcommand{\eqnref}[2][]{Equation#1~(\ref{eqn:#2})}

\newcommand\email{\begingroup \urlstyle{tt}\smaller\Url}

\title{End-to-end Network for Twitter Geolocation Prediction and 
Hashing}

\author{Jey Han Lau$^{1,2}$ \qquad Lianhua Chi$^{1}$ \qquad Khoi-Nguyen 
Tran$^{1}$ \qquad Trevor Cohn$^{2}$ \\[1ex]
    $^1$ IBM Research \\
    $^2$ School of Computing and Information Systems,\\The University of
Melbourne \\[1ex]
    \email{jeyhan.lau@gmail.com}, \email{lianhuac@au1.ibm.com}, \\
\email{khndtran@au1.ibm.com}, \email{t.cohn@unimelb.edu.au}}

\date{}

\begin{document}

\maketitle

\begin{abstract}

We propose an end-to-end neural network to predict the geolocation of a 
tweet.  The network takes as input 
a number of raw Twitter metadata such as the tweet message and 
associated user account information. Our model is language independent, 
and despite minimal feature engineering, it is interpretable and capable 
of learning location indicative words and timing patterns.  Compared to 
state-of-the-art systems, our model outperforms them by 2\%-6\%.  
Additionally, we propose extensions to the model to compress 
representation learnt by the network into binary codes. Experiments show 
that it produces compact codes compared to benchmark hashing algorithms.  
An implementation of the model is released 
publicly.\footnote{\url{https://github.com/jhlau/twitter-deepgeo}}

\end{abstract}

\section{Introduction}

A number of applications benefit from geographical information in 
social data, from personalised advertising to event detection to public 
health studies. \newcite{Sloan+:2013} estimate that less than 1\% of 
tweets are geotagged with their locations, motivating the development of 
geolocation prediction systems.

\newcite{Han+:2012c} introduced the task of predicting the location 
based only the tweet message. A key difference to previous work is that 
the prediction is made at message or tweet level, while predecessor 
methods tend to focus on user-level prediction
\cite{Backstrom+:2010,Cheng+:2010}. Since then, various methods have 
been proposed for the task 
\cite{Han+:2014a,Chi+:2016,Jayasinghe+:2016,Miura+:2016}, although most 
systems are engineered for a particular platform and language (e.g.\ 
website-specific parsers and language-specific tokenisers and 
gazetteers). Another strand of research leverages the social network 
structure to infer location; \newcite{Jurgens+:2015} provided a 
standardised comparison of these systems. Our focus in this paper is on 
using only the tweets, although \newcite{Rahimi+:2015a} showed that the 
best approach maybe to combine both types of information.

In applications where fast retrieval of co-located tweets is necessary 
(e.g.\ disaster detection), efficient representation of large volume of 
tweets constitute an important issue. 
Traditionally, hashing techniques such as locality sensitive hashing 
\cite{Indyk+:1998} are used to compress data into binary codes for fast 
retrieval (e.g.\ with multi-index hash tables 
\cite{Norouzi+:2012,Norouzi+:2014}), but it is not immediately clear how 
they can interact with raw Twitter metadata --- as they often require a 
vector as input --- and incorporate supervision.\footnote{In our case, 
the supervised information is the geolocation of tweets.}

To this end, we propose an end-to-end neural network for tweet-level 
geolocation prediction.  Our network is designed to be interpretable: we 
show it has the capacity to automatically learn location indicative 
words and activity patterns  from different regions.

Our contribution in this paper is two-fold.  First, our model 
outperforms state-of-the-art systems by 2-6\%, even though it has 
minimal feature engineering and is completely language-independent, as 
it uses no gazetteers or language preprocessing tools such as tokenisers 
or parsers. Second, our network can further compress learnt 
representations into compact binary code that incorporates information 
about the tweet and its geolocation.  To the best of our knowledge, this 
is the first end-to-end hashing method for tweets.


\section{Related Work}
\label{sec:related-work}

Early work in geolocation prediction operated at the user-level.  
\newcite{Backstrom+:2010} developed a methodology to predict the 
location of a user on Facebook by measuring the relationship between 
geography and friendship networks, and \newcite{Cheng+:2010} proposed a 
content-based prediction system to predict a Twitter  user's  location  
based  purely on his/her tweet messages. \newcite{Han+:2012c} introduced 
tweet-level prediction, where they first extract location indicative 
words by leveraging the geotagged tweets and then train a classifier for 
geolocation prediction using the location indicative words as features.  
Extending on this, systems were developed to better rank these location 
indicative or geospatial words by locality 
\cite{Chang+:2012,Laere+:2014,Han+:2014a}.  More recently, 
\newcite{Han+:2016} proposed a shared task for Twitter geolocation 
prediction, offering a benchmark dataset on the task.

Hashing is an effective method to compress data for fast access and 
analysis. Broadly there are two types of hashing techniques:   
data-independent techniques which design arbitrary functions to generate 
hashes, and data-dependent techniques that leverage pairwise similarity 
in the training data \cite{Chi+:2017}.  Locality-sensitive hashing 
(\lsh: \newcite{Indyk+:1998}) is a widely-known data-independent hashing 
method that uses randomised projections to generate hashcodes. It 
preserves data characteristics and guarantees the collision probability 
between data points.
\fh \cite{Lin+:2014}, on the other hand,  is a supervised data-dependent 
hashing that incorporates label information to determine pairwise 
similarity.  It uses decision tree based hash functions and graph 
cut-based binary code inference to deal with high dimensionality 
training data.

\begin{figure*}[t]
\begin{center}
\includegraphics[width=0.75\textwidth]{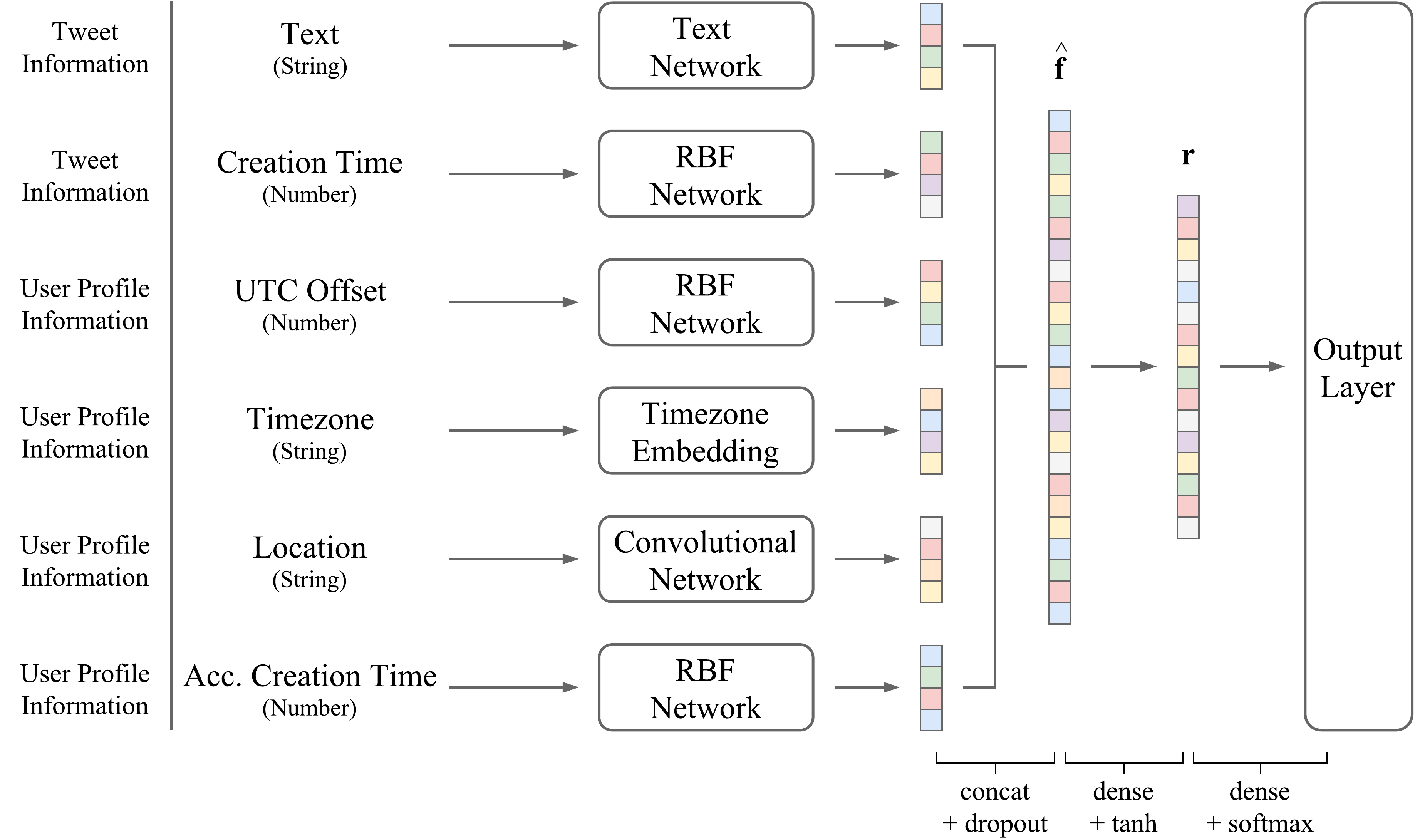}
\end{center}
\caption{Overall architecture of \deepgeo.}
\label{fig:architecture}
\end{figure*}

\section{Geolocation Prediction}

\subsection{Dataset}
\label{dataset}

We use the geolocation prediction shared task dataset \cite{Han+:2016} 
for our experiments. There are 2 proposed tasks, predicting geolocation 
given: (1) a tweet (tweet-level); and (2) a collection of tweets by a 
user (user-level).  For each task, there is a hard classification 
evaluation setting for predicting a city class, and a soft evaluation 
setting for predicting latitude and longitude coordinates.

We explore only the more challenging tweet-level prediction task.
In terms of evaluation setting, we experiment with the hard 
classification setting, where the network is required to predict one out 
of 3362 cities. Note that the metadata of a tweet includes not only the 
message but a variety of information such as creation time and user 
account data such as location and timezone.

Training, development and test partitions are provided by the shared 
task organisers.\footnote{The organisers provide full metadata for the 
test partition but only the tweet IDs for training and development 
partitions. We collect metadata for training/development tweets using 
the Twitter API.} We preprocess the data minimally, removing tweets that 
have less than 
5 characters in the training partition (development and testing data is 
  untouched) and keeping all character types that have occurred 
5 times or more in training. Unseen character tokens are represented by 
  $<$UNK$>$. Preprocessed statistics of the dataset is given in 
\tabref{dataset}.

\begin{table}[t]
\footnotesize
\begin{center}
\begin{tabular}{ccc}
\textbf{Partition} & \textbf{\#Tweets} & \textbf{\#Characters} \\
\hline
Training & 8.9M & 554M \\
Development & 7.2K & 439K \\
Test & 10K & 629K \\
\end{tabular}
\end{center}
\caption{Dataset statistics.}
\label{tab:dataset}
\end{table}

\begin{figure*}[t]
\begin{center}
\includegraphics[width=0.6\textwidth]{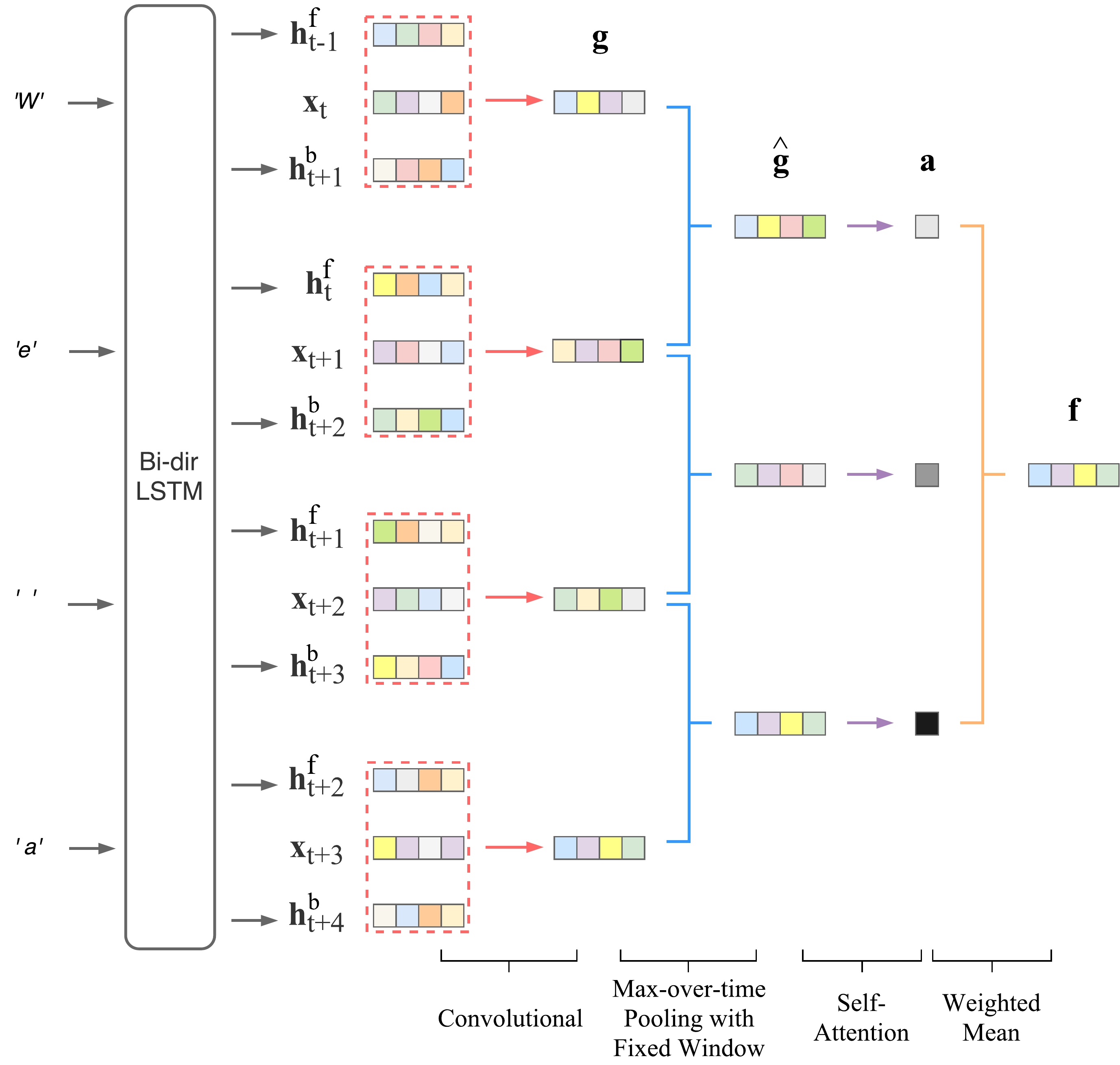}
\end{center}
\caption{Text Network.}
\label{fig:text-network}
\end{figure*}

\subsection{Network Architecture}

The overall architecture of our model (henceforth \deepgeo) is 
illustrated in \figref{architecture}. \deepgeo uses 6 features from the 
metadata: (1) tweet message; (2) tweet creation time; (3) user UTC 
offset; (4) user timezone; (5) user location; and (6) user account 
creation time.\footnote{We also tested user description and username, 
but preliminary experiments found these features are not very 
useful.}

Each feature from the metadata
is processed by a separate network to 
generate a feature vector $\mathbf{f}_j$. These feature vectors are then 
concatenated (with dropout applied) and connected to the penultimate 
layer:
\begin{align}
    \label{eqn:concat}
    \hat{\mathbf{f}} &= \mathbf{f}_1 \oplus \mathbf{f}_2 ... \oplus 
    \mathbf{f}_N \\
    \mathbf{r} &= \text{tanh}(\mathbf{W}_r \hat{\mathbf{f}})
\end{align}
where $N$ is the number of features (6 in total), $\mathbf{r} \in 
\mathbb{R}^{R}$ is the hidden representation at the penultimate layer 
and $\mathbf{W}_r$ is model parameter. For brevity, we omit biases in 
equations.

$\mathbf{r}$ is fully connected to the output layer and activated by 
softmax to generate a probability distribution over the classes. The 
model is trained with minibatches and optimised using Adam 
\citep{Kingma+:2014} with standard cross-entropy loss.

We design several networks for the raw features.  The first is a 
character-level recurrent convolutional network with a self-attention 
component for processing the tweet message (\secref{text-network}).  The 
second is an RBF network\footnote{Also known as mixture density 
network.} for processing numbers (\secref{rbf-network}).  The third is a 
simple convolutional network for processing user location 
(\secref{conv-network}), and the last is an embedding matrix for user 
timezone. We treat the timezone as a categorical feature, and learn 
embeddings for each timezone (309 unique timezones in total).  Note that 
these feature-processing networks are disjointed and there is no 
parameter sharing between them.

\subsubsection{Text Network}
\label{sec:text-network}

For the tweet message, we use a character-level recurrent convolutional 
network \cite{Lai+:2015}, followed by max-over-time pooling with a fixed 
window size and an attentional component to generate the feature vector, 
as illustrated in \figref{text-network}.

Let $\mathbf{x}_t \in \mathbb{R}^{E}$ denote the character embedding of 
the $t$-th character, we run a bi-directional LSTM network 
\cite{Hochreiter+:1997} to generate the forward and backward hidden 
states $\mathbf{h}^f_t$ and $\mathbf{h}^b_t$ respectively.\footnote{LSTM 
is implemented using one layer without any peep-hole connections and 
forget biases are initialised with 1.0.} 
We then concatenate the left and right context's hidden states with 
$\mathbf{x}_t$ and generate:
\begin{align*}
    \hat{\mathbf{x}}_t &= \mathbf{h}^f_{t-1} \oplus \mathbf{x}_t \oplus
\mathbf{h}^b_{t+1} \\
    \mathbf{g}_t &= \text{ReLU}(\mathbf{W}_g \hat{\mathbf{x}}_t) \\
\end{align*}
where $\hat{\mathbf{x}}_t \in \mathbb{R}^{3E}$, $\mathbf{W}_g \in 
\mathbb{R}^{O \times 3E}$ and $\mathbf{g}_t \in \mathbb{R}^{O}$. We 
iterate for each character to generate $\mathbf{g}_t$ for all time steps 
($\mathbf{W}_g$ can be interpreted as $O$ convolutional filters each 
with a window of $3 \times E$ striding 3 steps at a time).  Next, we 
apply max-over-time (narrow) pooling with window size $P$ over the 
vectors:
\begin{equation*}
    \hat{\mathbf{g}}_{t} = \text{max}(\mathbf{g}_t, \mathbf{g}_{t+1}, 
..., \mathbf{g}_{t+P-1})
\end{equation*}
where $\hat{\mathbf{g}}_{t} \in \mathbb{R}^{O}$ and max is a function 
that returns the element-wise maxima given a number of vectors of the 
same length. If there are $T$ characters in the tweet, this yields 
$(T-P+1)$ $\hat{\mathbf{g}}$ vectors, one for each span.

By setting $P = T$, we could generate one vector for the whole tweet.  
The idea of using a smaller window is that it enables a self-attention 
component, thereby allowing the network to discover the saliency of a 
character span --- for our task this means attending to location 
indicative words (\secref{qualitative-analyses}).  We define the 
attention network as follows:
\begin{align*}
    \alpha_{t} &= \mathbf{v}^\intercal \text{tanh}(\mathbf{W}_v 
\hat{\mathbf{g}}_{t}) \\
    \mathbf{a} &= \text{softmax}(\alpha_{0}, \alpha_{1}, ..., 
\alpha_{T-P})
\end{align*}
where $\mathbf{W}_v \in \mathbb{R}^{V \times O}$, $\mathbf{v} \in 
\mathbb{R}^{V}$ and $\mathbf{a} \in \mathbb{R}^{T-P+1}$. Given the 
attention, we compute a weighted mean to generate the final feature 
vector:
\begin{equation*}
    \mathbf{f}_{\text{text}} = \sum^{T-P}_{t=0} \mathbf{a}_t 
\hat{\mathbf{g}}_{t}
\end{equation*}
where $\mathbf{a}_t$ denotes the $t$-th element in $\mathbf{a}$, and 
$\mathbf{f}_{\text{text}} \in \mathbb{R}^{O}$.

\subsubsection{RBF Network}
\label{sec:rbf-network}


There are three time features in the metadata: tweet creation time, user 
account creation time and user UTC offset. The creation times are given 
in UTC time (i.e.\ not local time), e.g.\ \textit{Thu Jul 29 17:25:38 
+0000 2010} and the offset is an integer.

For the creation times, we use only time of the day information (e.g.\ 
\textit{17:25}) and normalise it from 
0 to 1.\footnote{As an example, \textit{17:25} is converted to 0.726.} 
  UTC offset is converted to hours and normalised to the same 
range.\footnote{UTC offset minimum is assumed -12 and maximum +14 based 
on: \url{https://en.wikipedia.org/wiki/List_of_UTC_time_offsets}.}

The aim of the network is to split time into multiple bins.  We can
interpret each hour as one bin (24 bins in total) 
and tweets originated from a particular location (e.g.\  Europe) favour 
certain hours or bins. This preference of bins should be different to 
tweets from a distant location (e.g.\ East Asia).  Assuming each bin 
follows a Gaussian distribution, then the goal of the network is to 
learn the Gaussian means and standard deviations of the bins.

Formally, given an input value $u$, for bin $i$ the network computes:
\begin{equation*}
    r_i = \exp \left( \frac{-(u - \mu_i)^2}{2\sigma_i^2} \right)
\end{equation*}
where $r_i$ is the output value and $\mu_i$ and $\sigma_i$ are the 
parameters for bin $i$. Let $B$ be the total number of bins, the feature 
vector generated by a RBF network is given as follows:
\begin{equation*}
    \mathbf{f}_{\text{rbf}} = [ r_0, r_1, ..., r_{B-1} ]
\end{equation*}
where $\mathbf{f}_{\text{rbf}} \in \mathbb{R}^B$.

\subsubsection{Convolutional Network}
\label{sec:conv-network}

Location is a user self-declared field in the metadata. As it is 
free-form text, we use a standard character-level convolution neural 
network \cite{Kim:2014} to process it. The network architecture is  
simpler compared to the text network (\secref{text-network}): it has no 
recurrent and self-attention layers, and max-over-time pooling is 
performed over all spans.

Let $\mathbf{x}_t \in \mathbb{R}^{E}$ denote the character embedding of 
the $t$-th character in the tweet. A tweet of $T$ characters is 
represented by a concatenation of its character vectors: 
$\mathbf{x}_{0:T-1} = \mathbf{x}_0 \oplus \mathbf{x}_1 \oplus ... \oplus 
\mathbf{x}_{T-1}$. We use convolutional filters and max-over-time 
(narrow) pooling to compute the feature vector:
\begin{align*}
    \mathbf{g}_t &= \text{ReLU}(\mathbf{W}_g \mathbf{x}_{t:t+Q-1})\\
   \mathbf{f}_{\text{conv}} &= \text{max}(\mathbf{g}_0, \mathbf{g}_{1}, 
..., \mathbf{g}_{T-Q})
\end{align*}
where $Q$ is the length of the character span, $\mathbf{W}_g \in 
\mathbb{R}^{O \times QE}$ ($\mathbf{W}_g$ can be interpreted as $O$ 
convolutional filters each with a window of $Q \times E$) and 
$\mathbf{g}_t, \mathbf{f}_{\text{conv}} \in \mathbb{R}^{O}$.

\subsection{Experiments and Results}

We explore two sets of features for predicting geolocation, using: (1) 
only the tweet message; and (2) both tweet and user metadata.  For the 
latter approach, we have 6 features in total (see 
\figref{architecture}). Classification accuracy is used as the metric 
for evaluation.

We tune network hyper-parameter values based on development accuracy;  
optimal hyper-parameter settings are presented in 
\tabref{hyper-parameters}. The column ``Message-Only'' uses only the 
text content of tweets, while ``Tweet$+$User'' incorporates both tweet 
and user account metadata.

For tweet message and user location, the maximum character length is set 
to 300 and 20 characters respectively; strings longer than this 
threshold are truncated and shorter ones are padded.\footnote{Tweets can  
exceed the standard 140-character limit due to the use of non-ASCII 
characters.} Models are trained using 10 epochs without early stopping.  
In each iteration, we reset the model's parameters if its development 
accuracy is worse than that of previous iteration.

We compare \deepgeo to 3 benchmark systems, all of which are systems 
submitted to the shared task \cite{Han+:2016}:

\begin{table}[t]
\begin{center}
\begin{adjustbox}{max width=0.4\textwidth}
\begin{tabular}{cccc}
\multirow{2}{*}{\textbf{Network}} & \textbf{Hyper-} & \textbf{Message-} 
&\textbf{Tweet$+$} \\
& \textbf{Parameter} & \textbf{Only} & \textbf{User} \\
\hline
\multirow{5}{*}{Overall} & Batch Size & \multicolumn{2}{c}{512} \\
& Epoch No. & \multicolumn{2}{c}{10}  \\
& Dropout & \multicolumn{2}{c}{0.2} \\
& Learning Rate & \multicolumn{2}{c}{0.001} \\
& $R$ & \multicolumn{2}{c}{400} \\
\hdashline
\multirow{3}{*}{Text} & Max Length & 300 & 300 \\
& $E$ & 200 & 200 \\
& $P$ & 10 & 10 \\
& $O$ & 600 & 400 \\
\hdashline
Time & $B$ & -- & 50 \\
\hdashline
UTC Offset & $B$ &  -- & 50 \\
\hdashline
\multirow{3}{*}{Timezone} & \multirow{2}{*}{Embedding} & 
\multirow{3}{*}{--} & \multirow{3}{*}{50}\\
& \multirow{2}{*}{Size} & & \\
& & & \\
\hdashline
\multirow{4}{*}{Location} & Max Length & -- & 20 \\
& $E$ & -- & 300 \\
& $Q$ & -- & 3 \\
& $O$ & -- & 300 \\
\hdashline
Account & \multirow{2}{*}{$B$} & \multirow{2}{*}{--} & 
\multirow{2}{*}{10} \\
Time & && \\
\end{tabular}
\end{adjustbox}
\end{center}
\caption{\deepgeo hyper-parameters and values.}
\label{tab:hyper-parameters}
\end{table}

\begin{table}[t]
\begin{center}
\begin{adjustbox}{max width=0.48\textwidth}
\begin{tabular}{ccc}

\textbf{Accuracy} & \textbf{System} & \textbf{Features} \\
\hline
0.146   & \newcite{Chi+:2016} & Message Only\\
\textbf{0.212} & \deepgeo & Message Only\\
\hdashline
0.409 & \newcite{Miura+:2016} & Tweet $+$ User Metadata\\
\textbf{0.428} & \deepgeo & Tweet $+$ User Metadata \\
\hdashline
\multirow{4}{*}{0.436} & \multirow{4}{*}{\newcite{Jayasinghe+:2016}} & 
Tweet $+$ User Metadata, \\
& & Gazetteer, \\
& & URL IP Lookup, \\
& & Label Prop. Network \\
\end{tabular}
\end{adjustbox}
\end{center}
\caption{Geolocation prediction test accuracy.}
\label{tab:classification-accuracy}
\end{table}

\begin{table*}[t]
\footnotesize
\begin{center}
\begin{adjustbox}{max width=1.0\textwidth}
\begin{tabular}{p{7cm}ccccc}
\multirow{2}{*}{\textbf{Tweet}} & \textbf{True} & \textbf{Predicted} & 
\multirow{2}{*}{\textbf{1st Span}} & \multirow{2}{*}{\textbf{2nd Span}} 
& \multirow{2}{*}{\textbf{3rd Span}} \\
& \textbf{Label} & \textbf{Label} &&& \\
\hline
\ex{Big thanks to @LouSnowPlow and all \#CleanSidewalk participants 
today.  You really make Louisville shine. To be happy, be 
compassionate!} &
\tl{3}{louisville-\\ky111-us} & \tl{3}{louisville-\\ky111-us} & 
\mn{3}{`Louisville'} & \mn{3}{`ake Louisv'} & \mn{3}{` Louisvill'} \\
\hdashline
\ex{McDonald's with aldha (@ Jalan A. P. Pettarani) 
http://t.co/HDVkhsKWBa} & \tl{2}{makassar-38-id} & 
\tl{2}{makassar-38-id} &
\mn{2}{`Pettarani)'} & \mn{2}{` Pettarani'} & \mn{2}{`ettarani) '} \\
\hdashline
\ex{Let's miss ALL the green lights on purpose! - every driver in 
Moncton this morning} & \tl{2}{moncton-04-ca} & \tl{2}{halifax-07-ca} & 
\mn{2}{`in Moncton'} & \mn{2}{`ncton this'} & \mn{2}{`Moncton th'} \\
\hdashline
\ex{Harrys bar toilet selfie @sophiethielmann @ Carluccio's Newcastle 
https://t.co/rKT7RGe7Nd} & \tl{2}{newcastle upon\\tyne-engi7-gb} & 
\tl{2}{newcastle upon\\tyne-engi7-gb} & \mn{2}{`s Newcastl'} & 
\mn{2}{`wcastle ht'} & \mn{2}{`castle htt'} \\
\hdashline
\ex{Makan terossssssss~ wkwkwk (with Erwina and Indah at McDonald's 
Bintara) - https://t.co/lT3KERFgap} & \tl{2}{bekasi-30-id} & 
\tl{2}{bekasi-30-id} & \mn{2}{` Bintara) '} & \mn{2}{`ntara) - h'} & 
\mn{2}{`intara) 
- '} \\
\hdashline
\ex{@EileenOttawa For better or worse it's a revenue stream for Twitter 
made available to businesses. We all have to get used to it.} & 
\tl{3}{toronto-08-ca} & \tl{3}{ottawa-08-ca} & \mn{3}{`tawa For b'} & 
\mn{3}{`ttawa For '} & \mn{3}{`nOttawa Fo'} \\
\hdashline
\ex{Hunt work!! (with @hadiseptiandani and @Febriantivivi at 
Jobforcareer Senayan) - https://t.co/u9myRDidtR} & \tl{2}{jakarta-04-id} 
& \tl{2}{jakarta-04-id} & \mn{2}{`nayan) - h'} & \mn{2}{` Senayan) '} & 
\mn{2}{`enayan) 
- '} \\
\end{tabular}
\end{adjustbox}
\end{center}
\caption{Examples of top character spans. Length of span is 
10 characters.  ``1st Span'' denotes the span that has the highest 
attention weight, ``2nd Span'' and ``3rd Span'' the second and third 
highest weight respectively.}
\label{tab:attention}
\end{table*}

\begin{table}[t]
\begin{center}
\begin{tabular}{lc@{\;}l}
\textbf{Feature Set} & \multicolumn{2}{c}{\textbf{Accuracy}} \\
\hline
All Features & 0.428 &  \\
\hdashline
$-$Text & 0.342 & {\small($-$0.086)} \\
$-$Tweet Creation Time & 0.419 & {\small($-$0.009)} \\
$-$UTC Offset & 0.431 & {\small($+$0.003)} \\
$-$Timezone& 0.422 & {\small($-$0.006)} \\
$-$Location& 0.228 & {\small($-$0.200)} \\
$-$Account Creation Time& 0.424 & {\small($-$0.004)} \\
\hline

\end{tabular}
\end{center}
\caption{Feature ablation results.}
\label{tab:ablation}
\end{table}

\begin{figure*}[t]
        \centering
        \begin{subfigure}[b]{0.3\textwidth}
                \includegraphics[width=\textwidth]{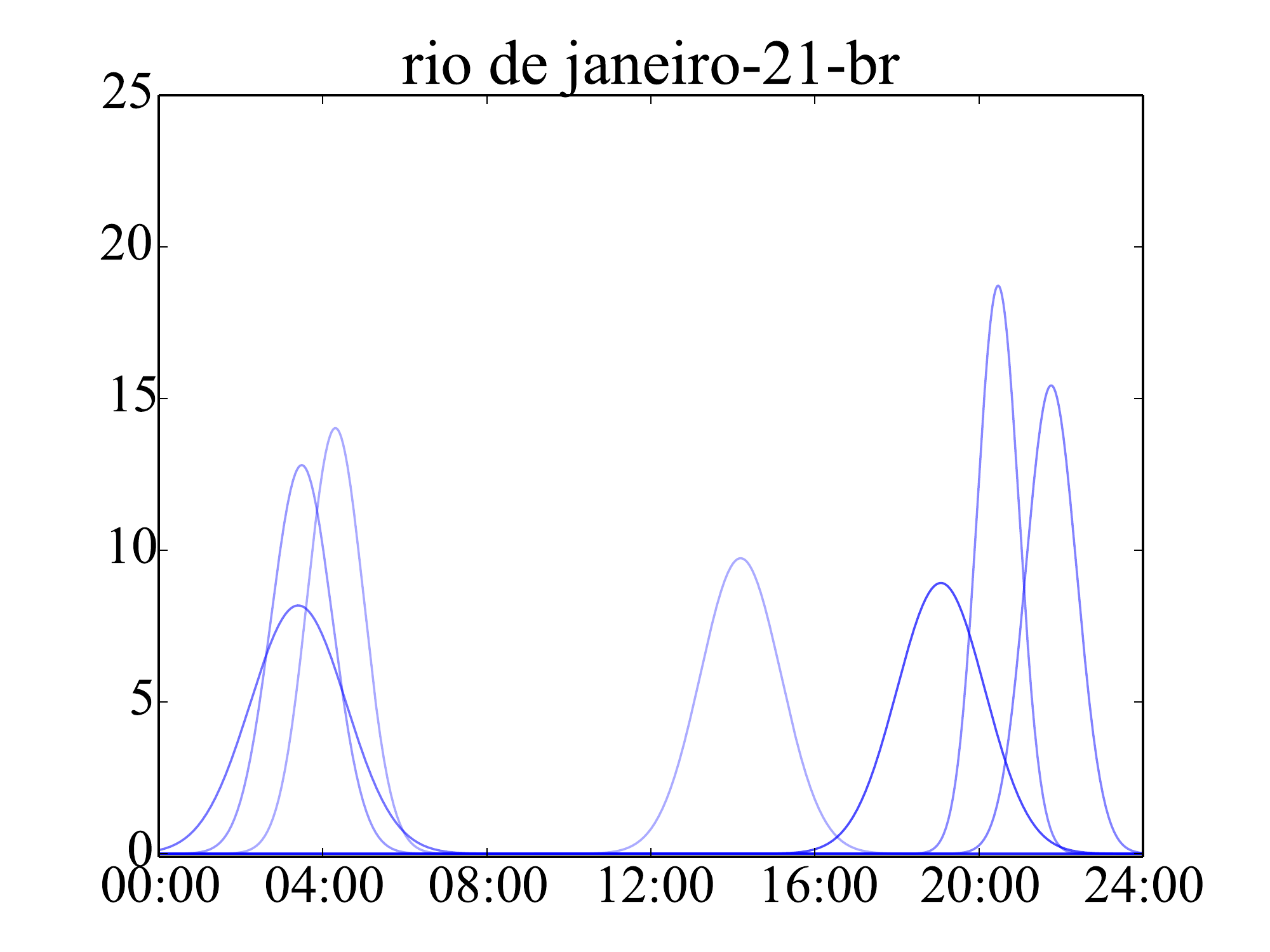}
                \caption{UTC-03}
                \label{fig:rio}
        \end{subfigure}
        ~
        \begin{subfigure}[b]{0.3\textwidth}
                \includegraphics[width=\textwidth]{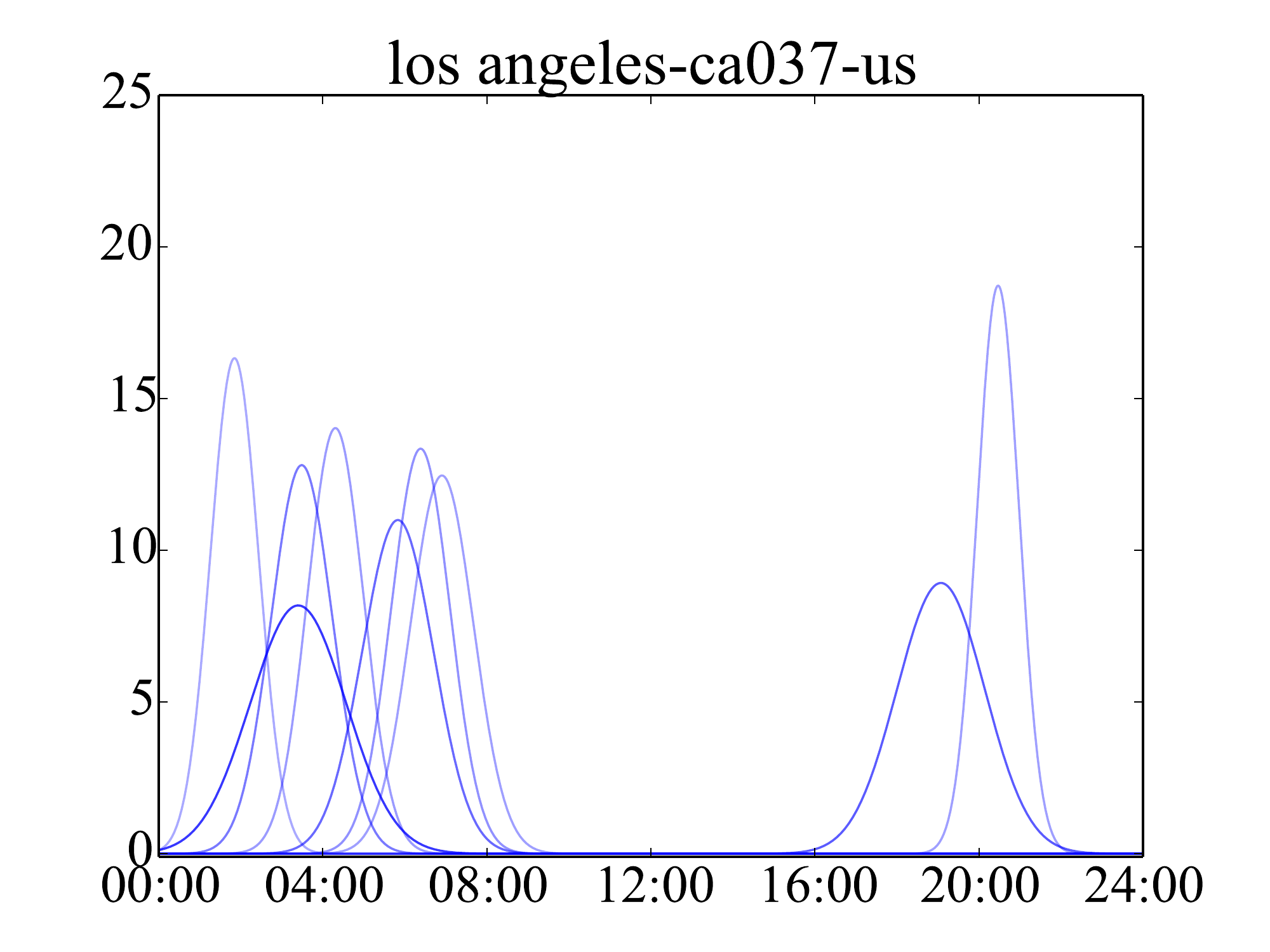}
                \caption{UTC-07}
                \label{fig:la}
        \end{subfigure}
        ~
        \begin{subfigure}[b]{0.3\textwidth}
                \includegraphics[width=\textwidth]{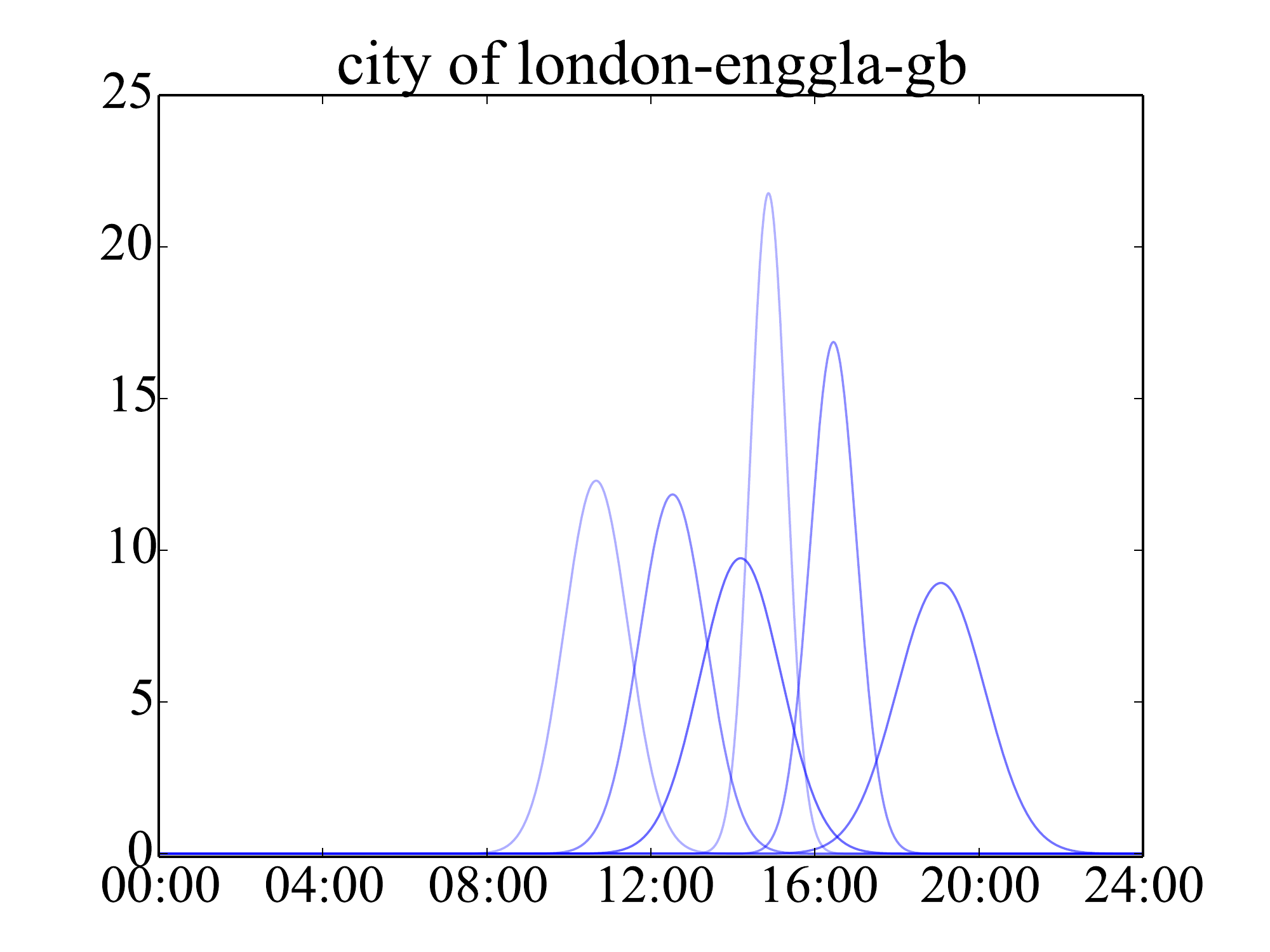}
                \caption{UTC+00}
                \label{fig:london}
        \end{subfigure}
        ~
        \begin{subfigure}[b]{0.3\textwidth}
                \includegraphics[width=\textwidth]{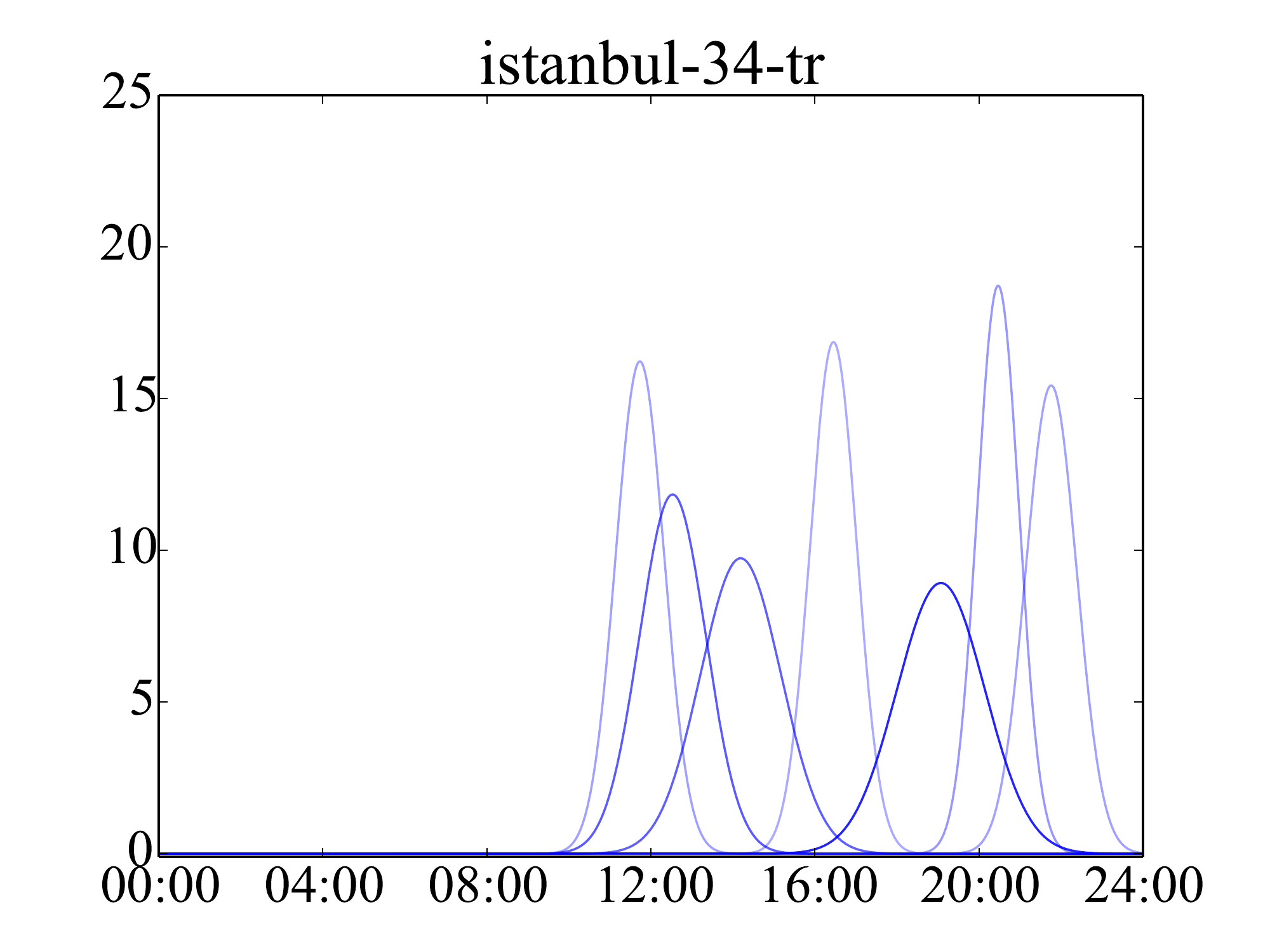}
                \caption{UTC+03}
                \label{fig:istanbul}
        \end{subfigure}
        ~
        \begin{subfigure}[b]{0.3\textwidth}
                \includegraphics[width=\textwidth]{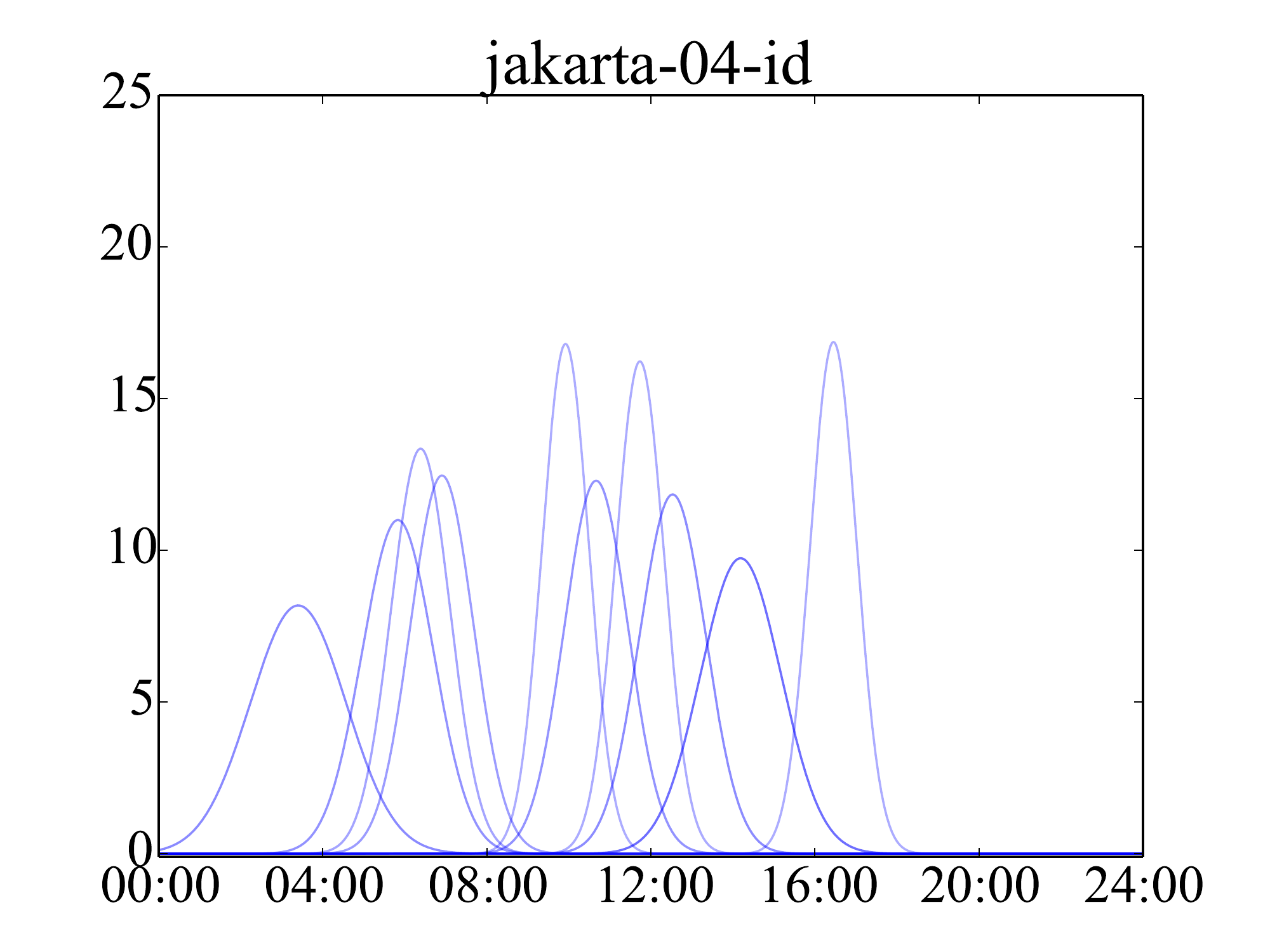}
                \caption{UTC+07}
                \label{fig:jakarta}
        \end{subfigure}
        ~
        \begin{subfigure}[b]{0.3\textwidth}
                \includegraphics[width=\textwidth]{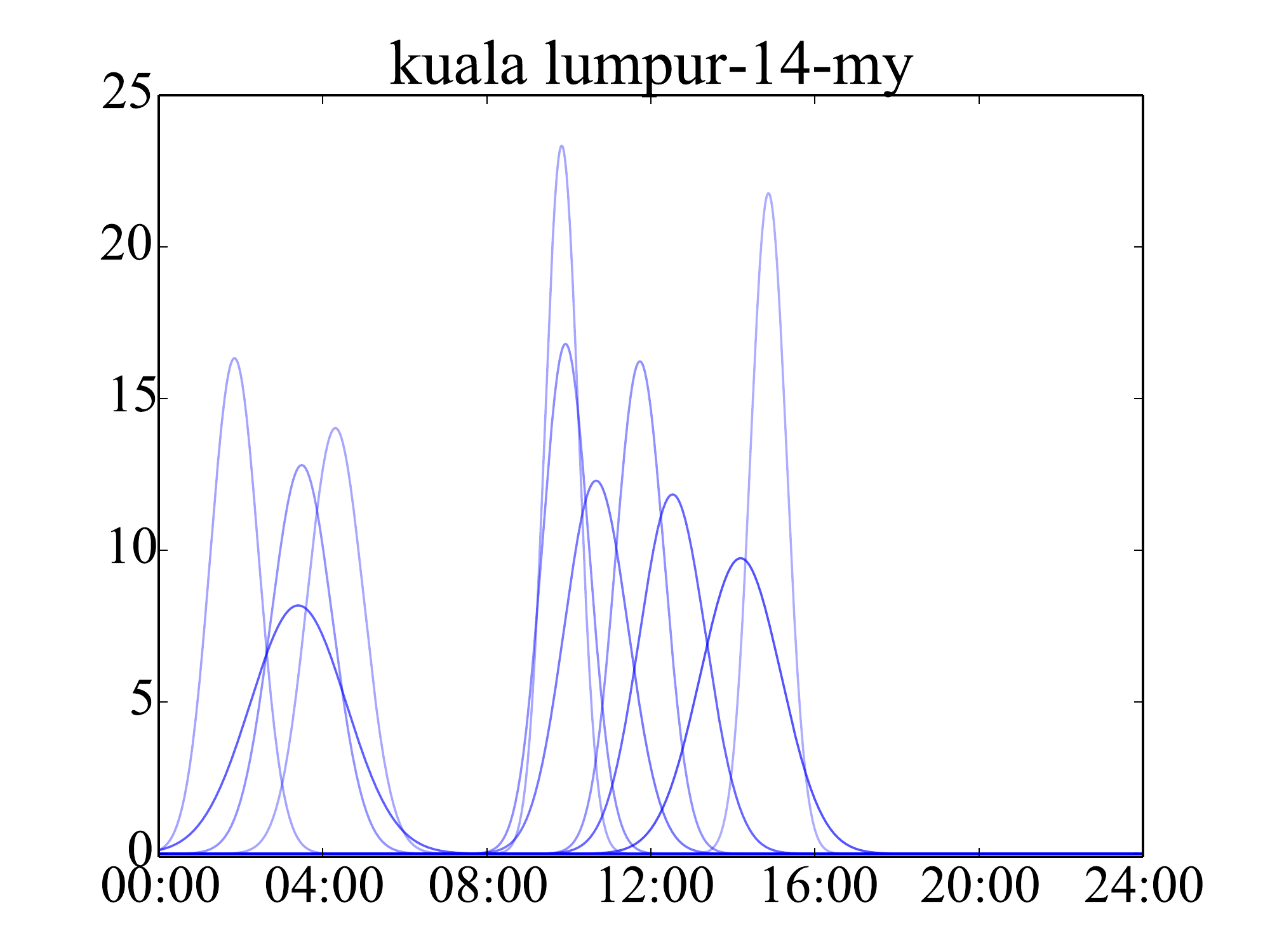}
                \caption{UTC+08}
                \label{fig:kl}
        \end{subfigure}
\caption{Tweet creation time distribution for 6 cities. Times in all 
plots are in UTC time.  Sub-caption indicates a city's UTC offset.}
\label{fig:time-distribution}
\end{figure*}

\paragraph*{\newcite{Chi+:2016}}
propose a geolocation prediction approach based on a multinomial naive 
Bayes classifier using a combination of automatically learnt location 
indicative words, city/country names, \#hashtags and @mentions.  
A frequency-based feature selection strategy is used to select the 
optimal subset of word features.

\paragraph*{\newcite{Miura+:2016}}
 experiment with a simple feedforward neural network for geolocation 
classification. The network draws inspiration from \fasttext 
\cite{Joulin+:2016}, where it uses mean word vectors to represent 
textual features and has only linear layers.  To incorporate multiple 
features --- tweet message, user location, user description and user 
timezone --- the network combines them via vector concatenation.

\paragraph*{\newcite{Jayasinghe+:2016}} develop an ensemble of 
classifiers for the task. Individual classifiers are built using a 
number of features indepedently from the metadata. In addition to using 
information embedded in the metadata, the system relies on external 
knowledgebases such as gazetteer and IP look up system to resolve URL 
links in the message. They also build a label propagation network that 
links connected users, as users from a sub-network are likely to come 
from the same location. These classifiers are aggregated via voting, and 
weights are manually adjusted based on development performance.

We present test accuracy performance for all systems in 
\tabref{classification-accuracy}.  Using only tweet message as feature, 
\deepgeo outperforms \newcite{Chi+:2016} by a considerable margin (over 
6\% improvement), even though  \deepgeo has minimal feature engineering 
and is trained at character level.
Next, we compare \deepgeo to \newcite{Miura+:2016}. Both systems use a 
similar set of features from the tweet and user metadata. \deepgeo sees 
an encouraging performance, with almost 2\% improvement. The best system 
in the shared task, \newcite{Jayasinghe+:2016}, remains the top 
performer. Note, however, that their system depends on language-specific 
processing tools (e.g.\ tokenisers), website-specific parsers (e.g.\ for 
extracting location information from user profile page on Instagram and 
Facebook) and external knowledge sources (e.g.\ gazetteers and IP 
lookup)  which were inaccessible by other systems.




To better understand the impact of each feature, we present ablation 
results where we remove one feature at a time in \tabref{ablation}. We 
see that the two most important features are the user location and tweet 
message.  These observations reveal that self-declared user location 
appears to be a reliable source of location, as task accuracy drops by 
almost half when this feature is excluded. For the other features, they 
generally have a small or negligible impact.

\begin{figure*}[t]
        \centering
        \begin{subfigure}[b]{0.15\textwidth}
                \includegraphics[width=\textwidth]{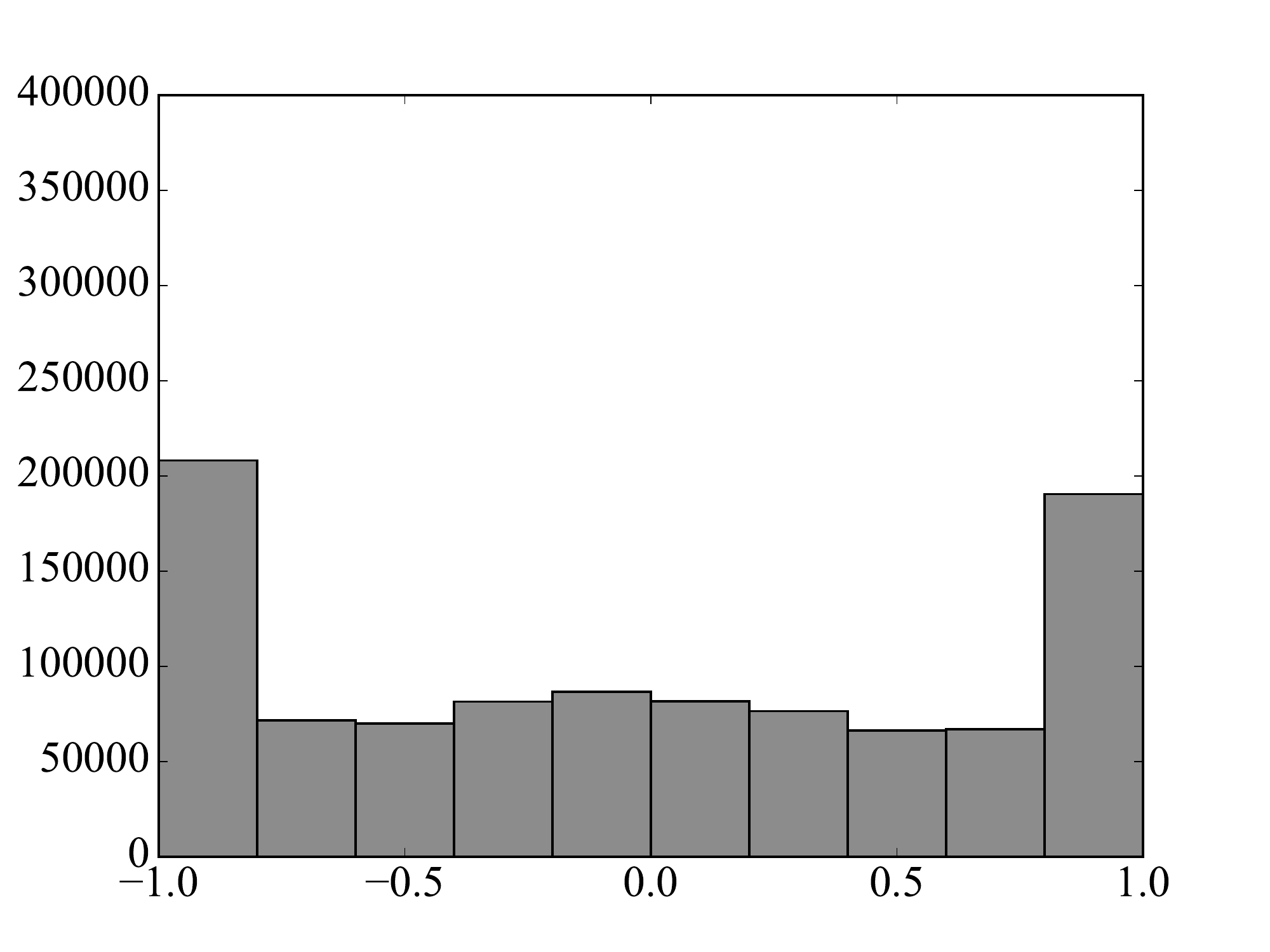}
                \caption{100; 0.0; 0.0}
        \end{subfigure}
        ~
        \begin{subfigure}[b]{0.15\textwidth}
                \includegraphics[width=\textwidth]{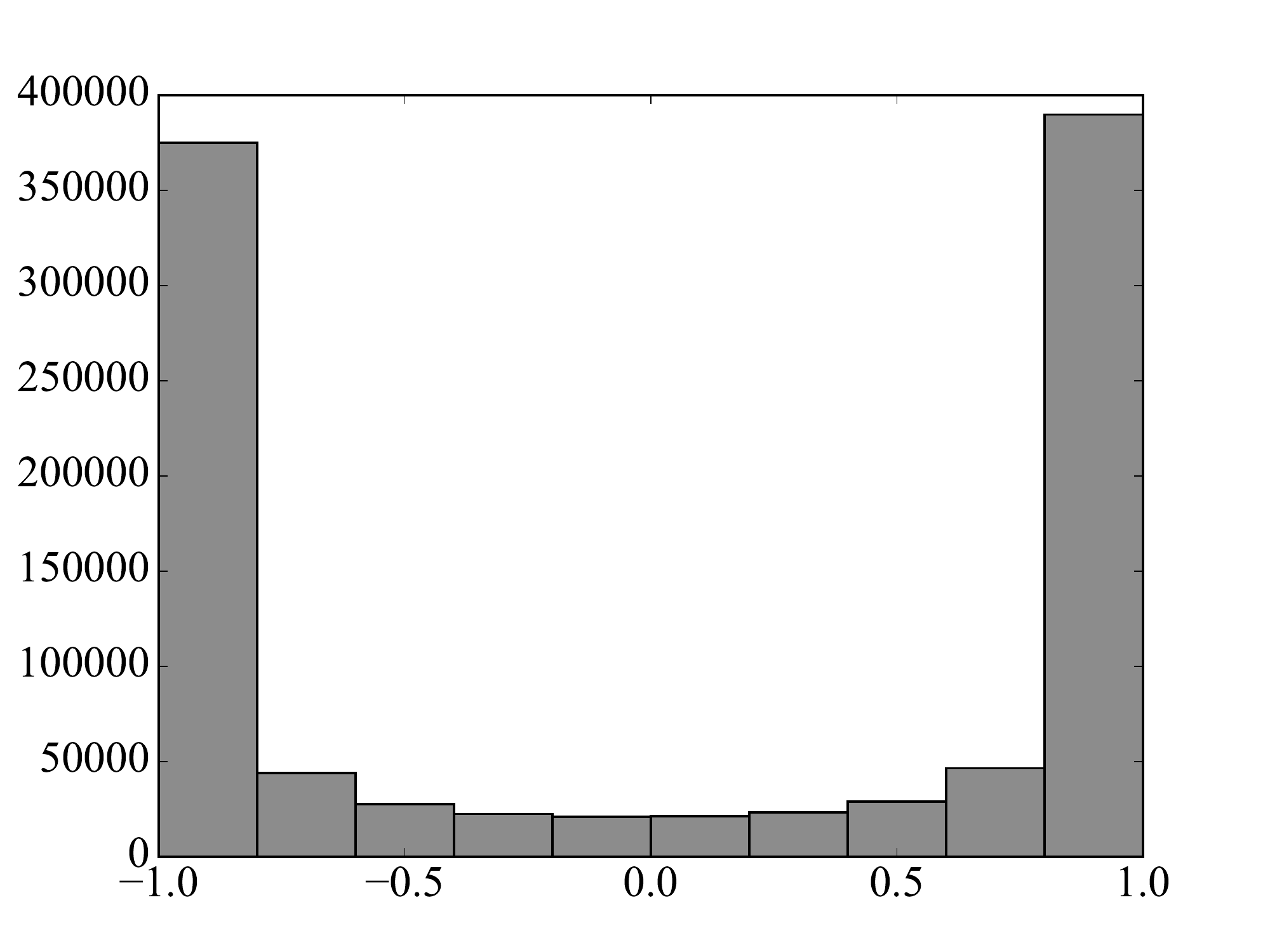}
                \caption{100; 0.1; 0.0}
        \end{subfigure}
        ~
        \begin{subfigure}[b]{0.15\textwidth}
                \includegraphics[width=\textwidth]{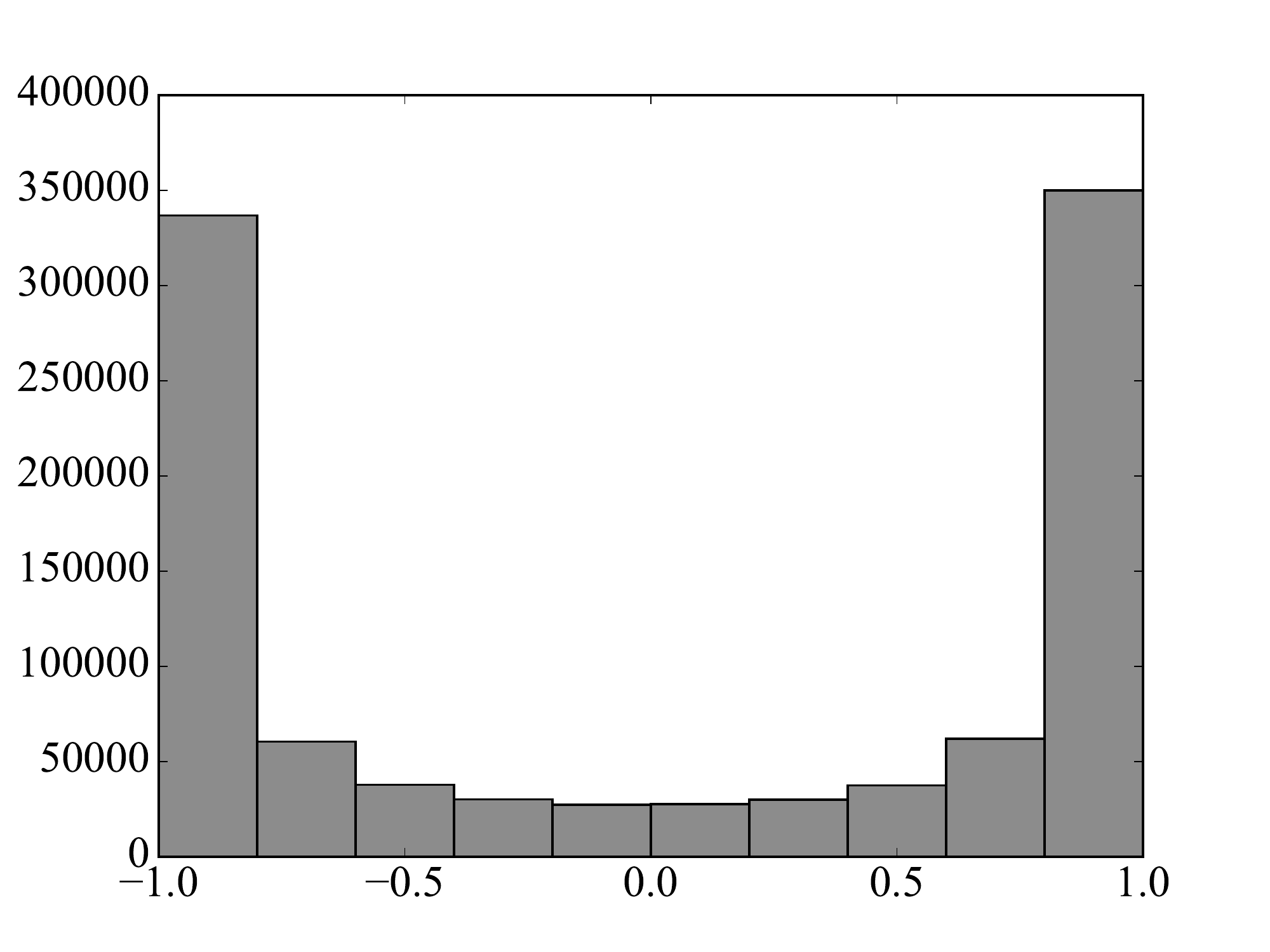}
                \caption{100; 0.0; 0.1}
        \end{subfigure}
        ~
        \begin{subfigure}[b]{0.15\textwidth}
                \includegraphics[width=\textwidth]{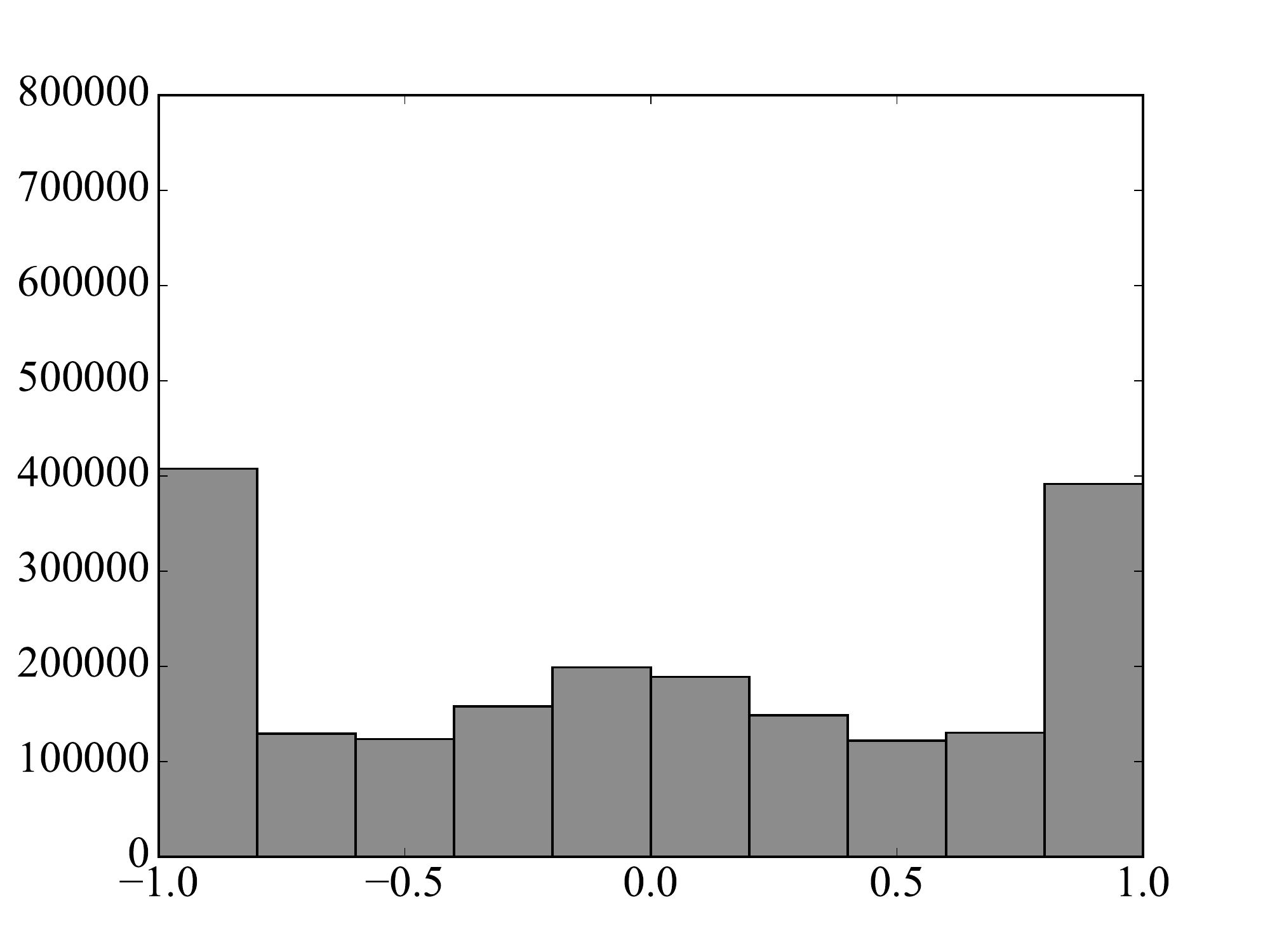}
                \caption{200; 0.0; 0.0}
        \end{subfigure}
        ~
        \begin{subfigure}[b]{0.15\textwidth}
                \includegraphics[width=\textwidth]{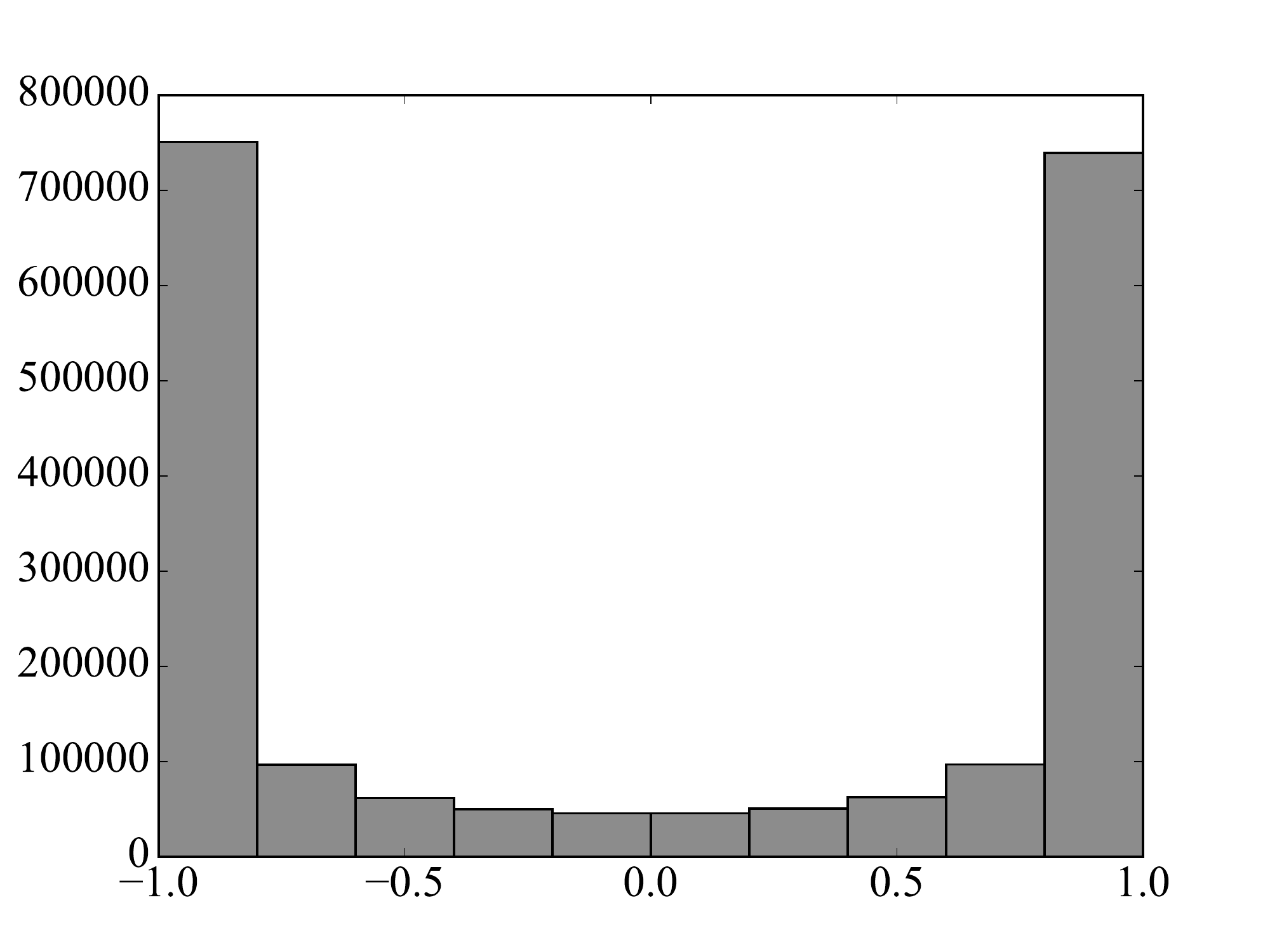}
                \caption{200; 0.1; 0.0}
        \end{subfigure}
        ~
        \begin{subfigure}[b]{0.15\textwidth}
                \includegraphics[width=\textwidth]{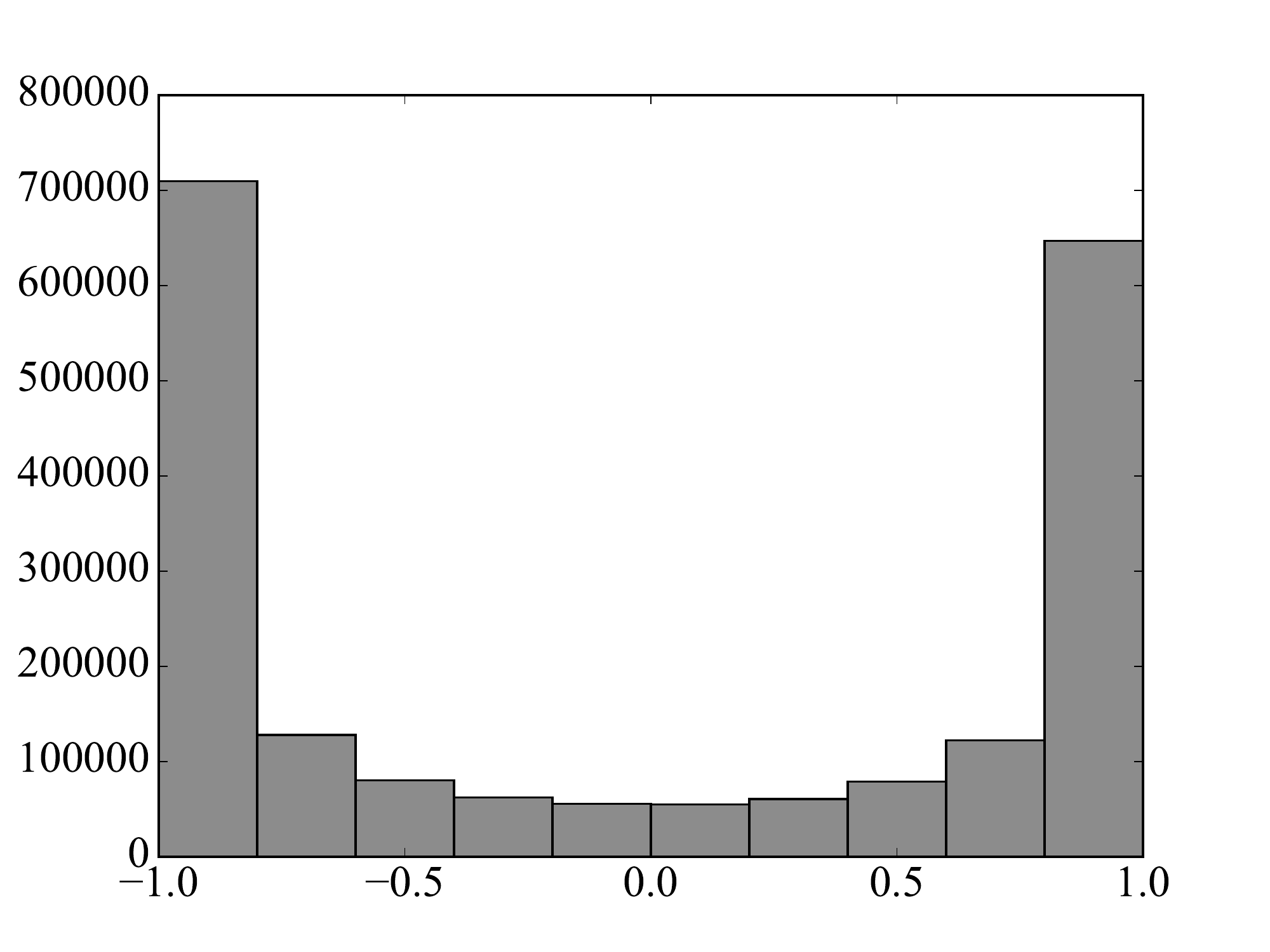}
                \caption{200; 0.0; 0.1}
        \end{subfigure}
        ~
        \begin{subfigure}[b]{0.15\textwidth}
                \includegraphics[width=\textwidth]{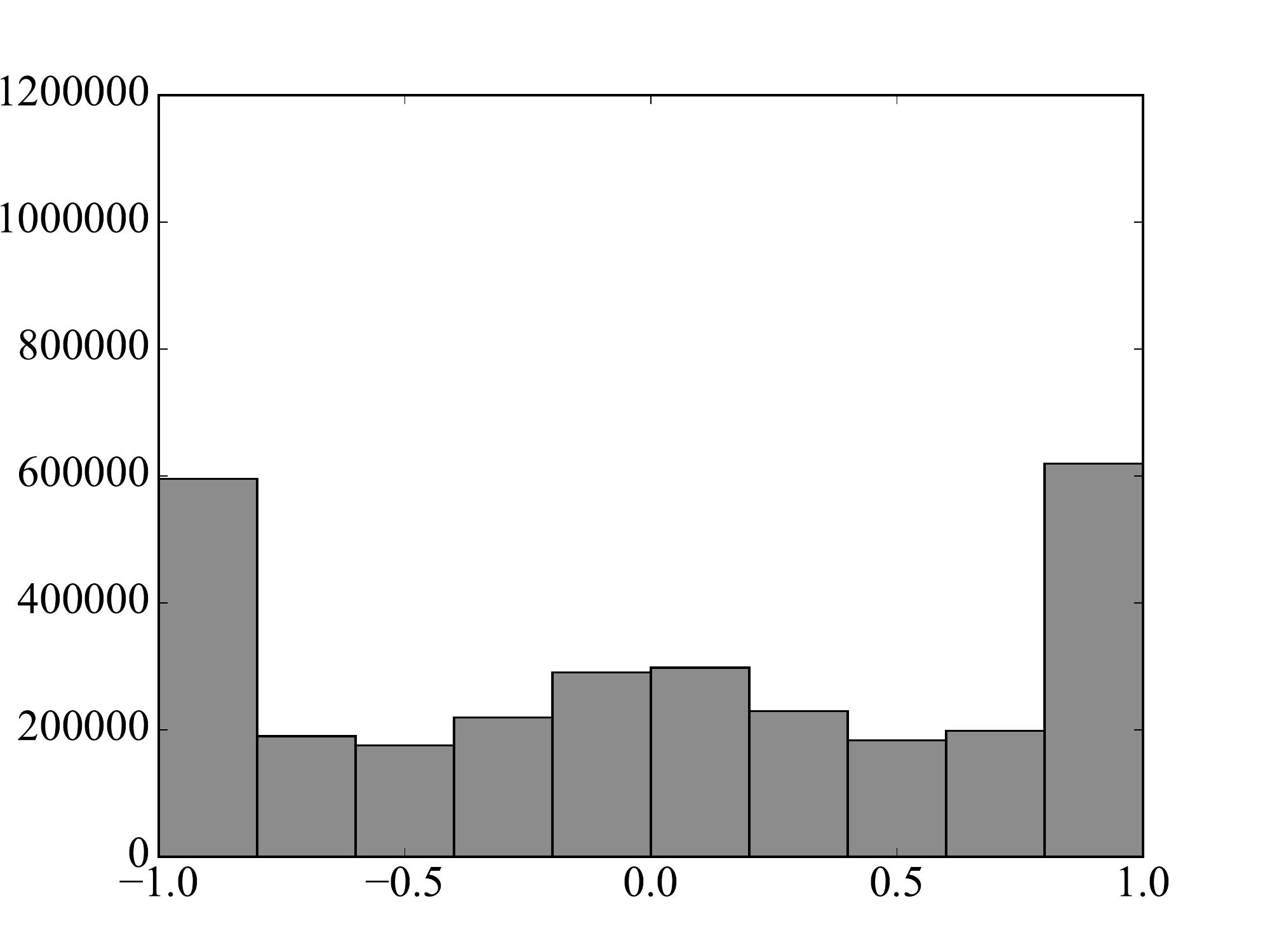}
                \caption{300; 0.0; 0.0}
        \end{subfigure}
        ~
        \begin{subfigure}[b]{0.15\textwidth}
                \includegraphics[width=\textwidth]{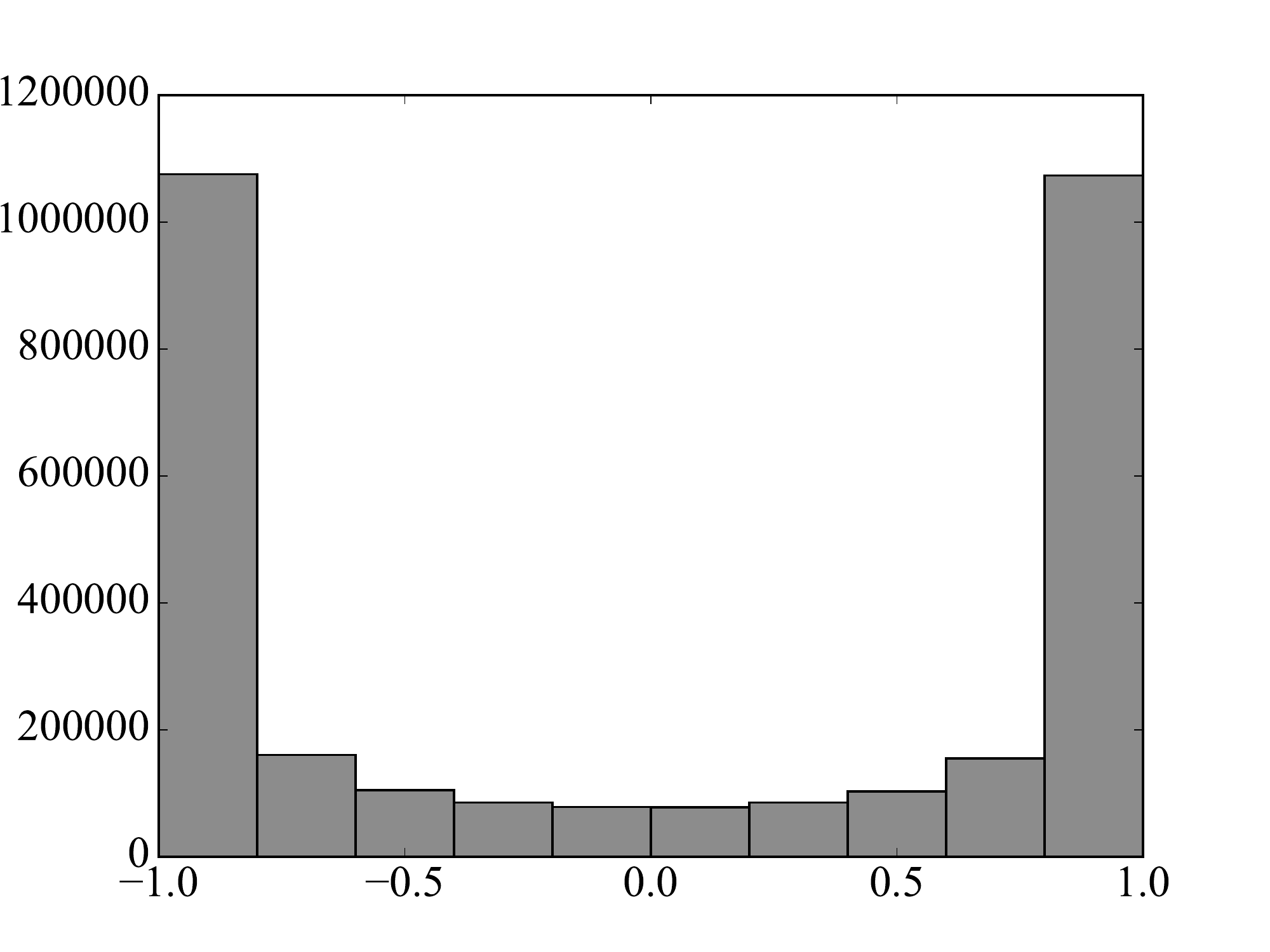}
                \caption{300; 0.1; 0.0}
        \end{subfigure}
        ~
        \begin{subfigure}[b]{0.15\textwidth}
                \includegraphics[width=\textwidth]{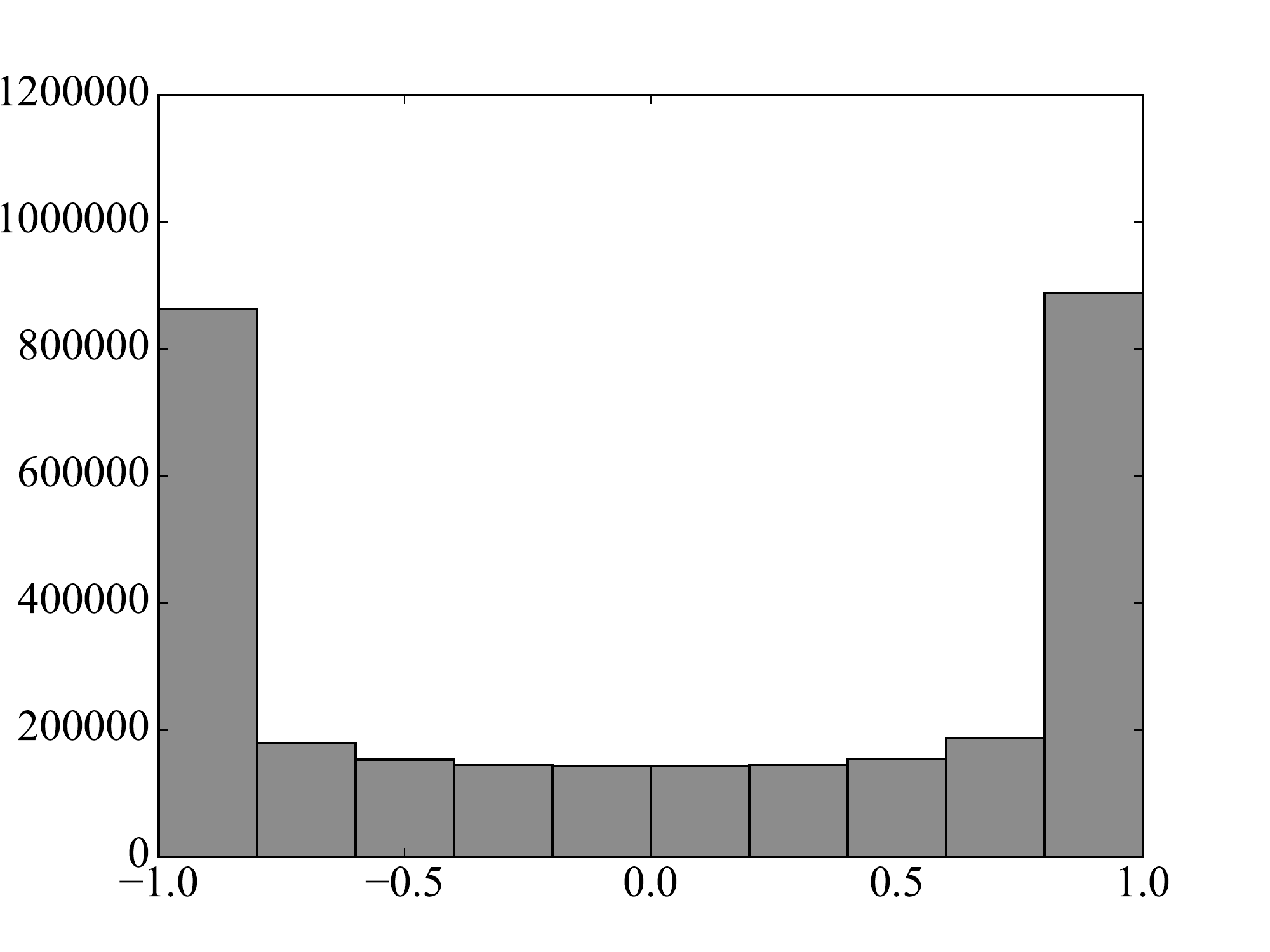}
                \caption{300; 0.0; 0.1}
        \end{subfigure}
        ~
        \begin{subfigure}[b]{0.15\textwidth}
                \includegraphics[width=\textwidth]{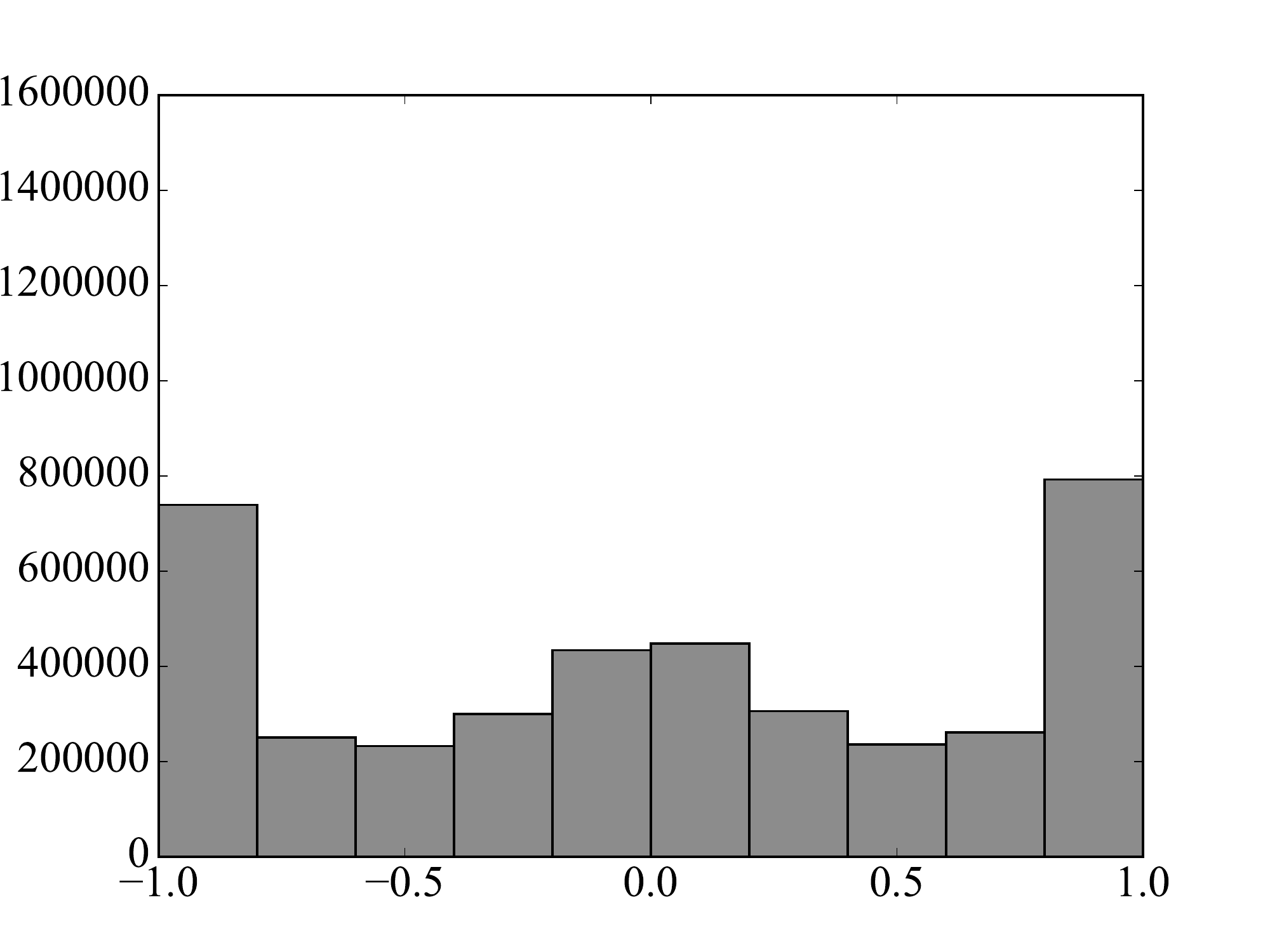}
                \caption{400; 0.0; 0.0}
        \end{subfigure}
        ~
        \begin{subfigure}[b]{0.15\textwidth}
                \includegraphics[width=\textwidth]{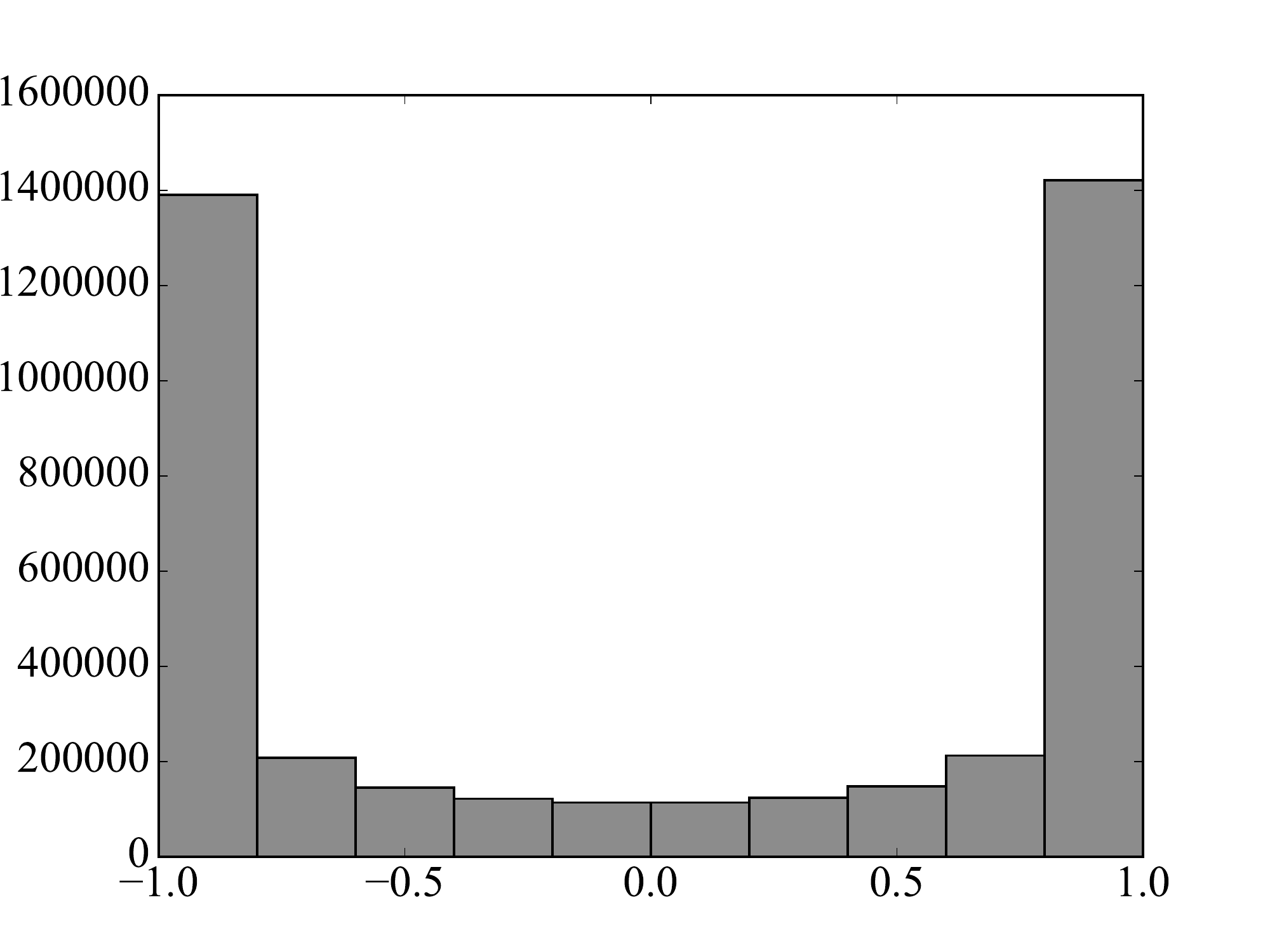}
                \caption{400; 0.1; 0.0}
        \end{subfigure}
        ~
        \begin{subfigure}[b]{0.15\textwidth}
                \includegraphics[width=\textwidth]{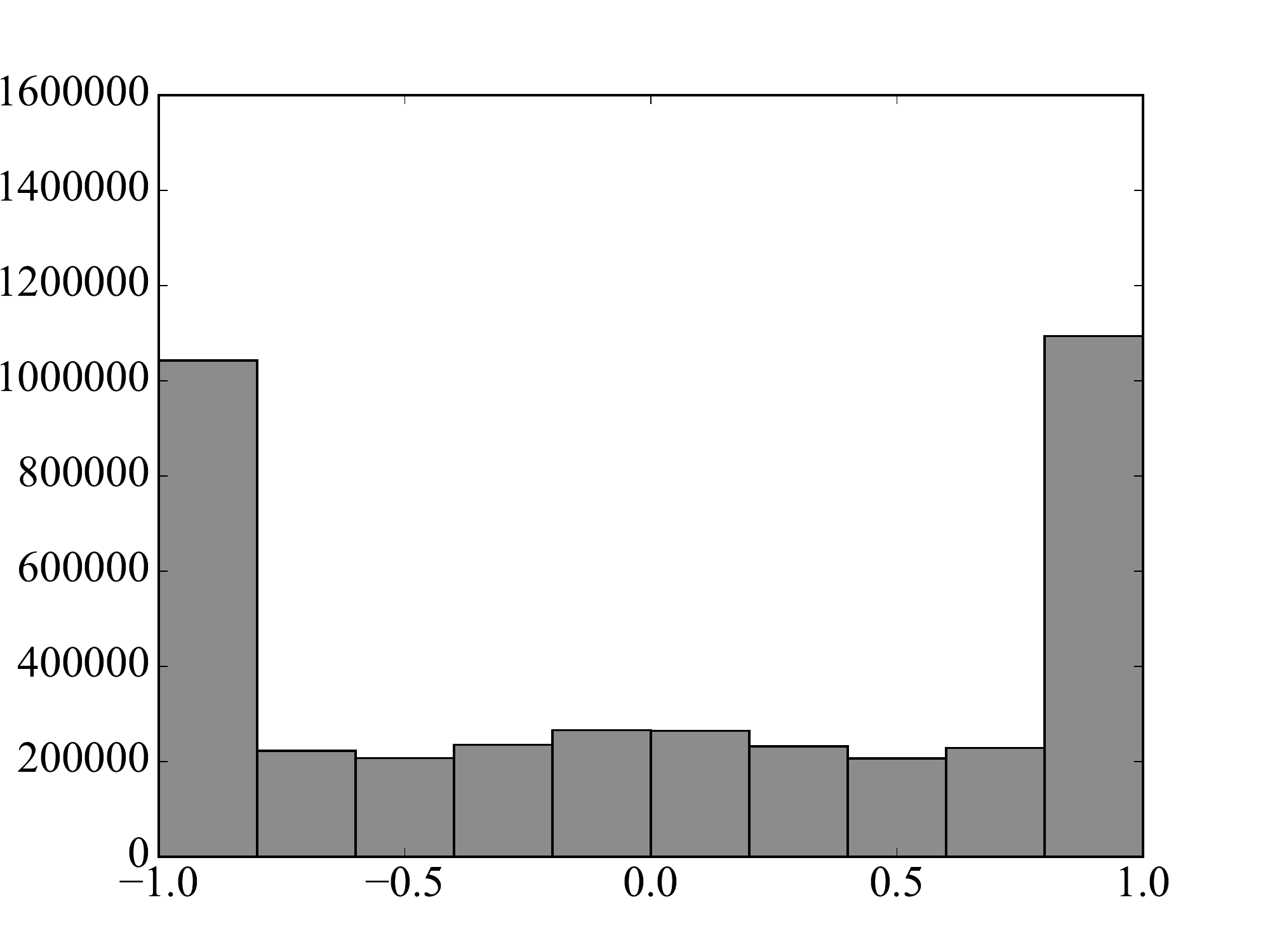}
                \caption{400; 0.0; 0.1}
        \end{subfigure}
\caption{Histogram of $\mathbf{r}$ element values. Fields in the 
subcaptions denote vector dimension ($R$), corruption level ($\sigma$) 
and $l$ scaling factor ($\alpha$). Range of y-axis for the same vector 
dimension is standardised.}
\label{fig:denoise}
\end{figure*}

\subsection{Qualitative Analyses}
\label{sec:qualitative-analyses}

The self-attention component in the text network 
(\secref{text-network})  captures saliency of character spans. To 
demonstrate its effectiveness, we select a number of tweets from the 
test partition and present the top-3 spans that have the highest 
attention weights in \tabref{attention}.

Interestingly, we see that whenever a location word is in the message, 
\deepgeo tends to focus around it (e.g.\ \textit{Louisville} and 
\textit{Newcastle}). Occasionally this can induce error in prediction, 
e.g.\ in the second last example the network focuses on \textit{Ottawa} 
even though the word has little significance to the true location 
(\textit{Toronto}).  Focussing on the right span does not necessarily 
result in correct prediction as well; as we see in the third example the 
network focuses on \textit{Moncton} but predicts a neighbouring city 
\textit{Halifax} as the geolocation.

Next, we look at the Gaussian mixtures learnt by the RBF network.  Using 
the gold-standard city labels, we collect
bin weights ($\mathbf{f}_{\text{rbf}}$) for tweet creation times (from 
test data) for  6 cities and plot them in \figref{time-distribution}.  
Each Gaussian distribution represents one bin, and its weight is 
computed as a mean weight over all tweets belonging to the city (line 
transparency indicates bin weight). Bins that have a mean weight $< 
0.075$ are excluded.

  For London (\figref{london}), we 
see that most tweets are created from 
10:00--20:00 local time.\footnote{London's UTC offset is +00 so no 
   adjustment is necessary to convert to local time.}  For Jakarta 
(\figref{jakarta}), tweet activity mostly centers around 
11:00--23:00 local time (04:00--16:00 UTC time). Most cities share a 
   similar activity period, with the exception of Istanbul 
(\figref{istanbul}):  Turkish people seems to start their day much 
later, as tweets begin to appear from 
15:00--01:00 local time (12:00 to 22:00 UTC time).

Another interesting trend we find is that for two cities (Los Angeles 
and Kuala Lumpur), there is a brief period of inactivity around noon 
(12:00) to evening (18:00) in local time. We hypothesise that most 
people are working during these times, and are thus too busy to use Twitter.

\section{Hashing: Generating Binary Code For Tweets}

\begin{table*}[t]
\begin{center}
\begin{adjustbox}{max width=0.8\textwidth}
\begin{tabular}{c:ccc:cc:cc}
\multirow{2}{*}{\textbf{Bits}} &
\multirow{2}{*}{\textbf{\deepgeo}} &
\textbf{\deepgeo} &
\textbf{\deepgeo} &
\multicolumn{2}{c:}{\textbf{\lsh}} &
\multicolumn{2}{c}{\textbf{\fh}} \\
&&\textbf{\method{$+$noise}} & \textbf{\method{$+$loss}} &  \textbf{\wv} 
& \textbf{\deepgeo} &
\textbf{\wv} & \textbf{\deepgeo} \\
\hline
100 & 0.147 & \textbf{0.149} & 0.146 & 0.013 & 0.053 & 0.116 & 0.140 \\
200 & 0.143 & 0.143 & 0.140 & 0.019 & 0.072 & 0.128 & \textbf{0.160} \\
300 & 0.136 & 0.137 & 0.141 & 0.021 & 0.082 & 0.133 & \textbf{0.165} \\
400 & 0.132 & 0.135 & 0.136 & 0.022 & 0.086 & 0.135 & \textbf{0.170} \\
\end{tabular}
\end{adjustbox}
\end{center}
\caption{Retrieval MAP performance.}
\label{tab:map}
\end{table*}

\deepgeo creates a low-dimensional dense vector representation  
($\mathbf{r}$) for a tweet in the penultimate layer. This representation 
captures the message, user timezone and other metadata (including the 
city label) that are incorporated to the network during training.

Storing the dense vector representation for a large volume of tweets can 
be costly.\footnote{If a tweet is represented by a vector of 
400 32-bit floating point numbers, 1B tweets would take 1.6TiB of 
space.} If we can compress the dense vectors into compact binary codes, 
it would save storage space, as well as enabling more efficient 
retrieval of co-located tweets, e.g.\ using multi-index hash tables for 
$K$-nearest neighbour search \cite{Norouzi+:2012,Norouzi+:2014}.


Inspired by denoising autoencoders \cite{Yu+:2016,Vincent+:2010}, we 
\textit{binarise} the dense vector generated by \deepgeo by adding 
Gaussian noise. The intuition is that the addition of noise sharpens the 
activation values in order to counteract the random noise.

\eqnref{concat} is thus modified to:
    $\hat{\mathbf{f}} = (\mathbf{f}_1 \oplus \mathbf{f}_2 ... \oplus 
    \mathbf{f}_N) + \mathcal{N}(0, \sigma^2)$, 
where $\mathcal{N}(0,\sigma^2)$ is a zero-mean Gaussian noise with 
standard deviation (or corruption level) $\sigma$.\footnote{Dropout is 
applied to $\hat{\mathbf{f}}$, i.e.\ after the addition of noise.} 

\begin{table}[t]
\begin{center}
\begin{adjustbox}{max width=0.4\textwidth}
\begin{tabular}{cccc}

\multirow{2}{*}{\textbf{$R$}} & \multirow{2}{*}{\textbf{\deepgeo}} &  
\textbf{\deepgeo} & \textbf{\deepgeo} \\
& & \textbf{$+$noise} & \textbf{$+$loss} \\
\hline
100 & \textbf{0.420} & 0.396 & 0.410 \\
200 & \textbf{0.428} & 0.417 & 0.414 \\
300 & \textbf{0.420} & 0.416 & 0.422 \\
400 & \textbf{0.428} & 0.419 & 0.418 \\

\end{tabular}
\end{adjustbox}
\end{center}
\caption{Classification performance for \deepgeo with the addition of 
noise and loss term $l$.}
\label{tab:classification-noise}
\end{table}

In the addition to the Gaussian noise, we also experiment with an 
additional loss term $l$ to penalise elements that are not in the 
extrema:
$l = \alpha \times \frac{1}{R} \sum_{i=0}^{R-1} 
|(\mathbf{r}_i-1)(\mathbf{r}_i+1)|$, 
where $\mathbf{r}_i$ is the $i$-th element in $\mathbf{r}$ and $\alpha$ 
is a scaling factor.  We set $\sigma=\alpha=0.1$, as both values were 
found to provide good performance.

To better understand the effectiveness of the noise and loss term $l$ in 
binarising the vector values, we present a histogram plot of 
$\mathbf{r}$ element values from test data in \figref{denoise}, for $R = 
100, 200, 300, 400$. We see that the addition of noise and $l$ helps in 
pushing the elements to the extrema. The noise term appears to work a 
little better than $l$, as the frequency for the $-1.0$ and $+1.0$ bins 
is higher. We also observe that there is a small increase in middle/zero 
values as R increases from 100 to 400, suggesting that there are more 
unused hidden units when number of parameters increases. We present 
classification accuracy performance when we add noise (\deepgeon) and 
$l$ (\deepgeol) in \tabref{classification-noise}. The  performance drops 
a little, but generally stays within a gap of 1\%.  This suggests that 
both noise and $l$ works well in binarising $\mathbf{r}$ without trading 
off classification accuracy significantly.

Next we evaluate the retrieval performance using the binary codes. We 
binarise $\mathbf{r}$ for development and test tweets using the sign 
function.
Given a test tweet, we retrieve the nearest development tweets based on 
hamming distance, and calculate average precision.\footnote{We remove 
1328 test tweets that do not share city labels with any development 
     tweets.
} We aggregate the retrieval performance for all test tweets by 
computing mean average precision (MAP).

For comparison, we experiment with two hashing techniques: \lsh 
\cite{Indyk+:1998} and \fh \cite{Lin+:2014} (see \secref{related-work} 
for system descriptions). The input required for both \lsh and \fh is a 
vector. We test 2 types of input for these methods: (1) a \wv baseline, 
where we concatenate mean word vectors of the tweet message, user 
account's timezone and location, resulting in a $900$-dimension 
vector;\footnote{$300$-dimension \wv (\sg) vectors are trained on 
English Wikipedia.} and (2) \deepgeo representation $\mathbf{r}$. The 
rationale for using \deepgeo as input is to test whether its 
representation can be further compressed with these hashing 
techniques.\footnote{\lsh and \fh are trained using 400K tweets due to 
large memory requirement. We also tested these models using only 150K 
tweets, and found marginal performance improvement from 150K to 400K, 
suggesting that they are unlikely to improve even if it is trained with 
the full data.}

We present MAP performance for all systems in \tabref{map}. Looking at 
\deepgeo systems (column 2--4), we see that adding noise and $l$ helps, 
although the impact is greater when the bit size is large (300/400 
bits). For \lsh, which uses no label information, \wv input produces 
poor binary code for retrieval.  Changing the input to \deepgeo improves 
retrieval considerably, implying that the representation produced by 
\deepgeo captures geolocation information.

\fh with \wv input vector performs competitively.  For smaller bit sizes 
(100 or 200), however, the gap in performance is substantial.  Pairing 
\fh with \deepgeo produces the best retrieval performance: for 
200/300/400 bits it outperforms \deepgeon by 2--4\%.  Interestingly for 
100 bits \fh is unable to compress \deepgeo's representation any 
    further, highlighting the compactness of \deepgeo representation for 
smaller bit sizes.


%
%

\section{Conclusion}

We propose an end-to-end method for tweet-level geolocation prediction.  
We found strong performance, outperforming comparable systems by 2-6\% 
depending on the feature setting. Our model is generic and has minimal 
feature engineering, and as such is highly portable to problems in other 
domains/languages (e.g.\ Weibo, a Chinese social platform, is one we 
intend to explore).  We propose simple extensions to the model to 
compress the representation learnt by the network into binary codes.  
Experiments demonstrate its compression power compared to 
state-of-the-art hashing techniques.


\bibliography{bibs/papers,bibs/ref}
\bibliographystyle{ijcnlp2017}

\end{document}